\documentclass{article} 
\usepackage{iclr2027_conference,times}
\def\cast{{
   \mathord{
      \hbox to 0em{
         \ooalign{
	   \smash{\hbox{$\ast$}}\crcr
	   \smash{\hskip-1pt\Large\hbox{$\circ$}} }
	 \hidewidth}
      \phantom{\bigcirc}
} }}

\newcommand{\bds}{\begin {itemize}}
\newcommand{\eds}{\end {itemize}}
\newcommand{\bdf}{\begin{definition}}
\newcommand{\blm}{\begin{lemma}}
\newcommand{\edf}{\end{definition}}
\newcommand{\elm}{\end{lemma}}
\newcommand{\bthm}{\begin{theorem}}
\newcommand{\ethm}{\end{theorem}}
\newcommand{\bprp}{\begin{prop}}
\newcommand{\eprp}{\end{prop}}
\newcommand{\bcl}{\begin{claim}}
\newcommand{\ecl}{\end{claim}}
\newcommand{\bcr}{\begin{coro}}
\newcommand{\ecr}{\end{coro}}
\newcommand{\bquest}{\begin{question}}
\newcommand{\equest}{\end{question}}

\newcommand{\larrow}{{\larrow}}

\newcommand{\argmin}{\ensuremath{\mathrm{arg}\min}}
\newcommand{\argmax}{\ensuremath{\mathrm{arg}\max}}

\newcommand{\cI}{{\ensuremath{\mathcal{I}}}}

\newcommand{\cL}{{\ensuremath{\mathcal{L}}}}

\newcommand{\cS}{{\ensuremath{\mathcal{S}}}}

\newcommand{\cW}{{\ensuremath{\mathcal{W}}}}

\newcommand{\mB}{{\ensuremath{\mathbf{B}}}}

\newcommand{\mM}{{\ensuremath{\mathbf{M}}}}

\newcommand{\mP}{{\ensuremath{\mathbf{P}}}}

\newcommand{\mX}{{\ensuremath{\mathbf{X}}}}

\usepackage{latexsym}

\def\IC{\mathbb C}
\def\IN{\mathbb N}
\def\IZ{\mathbb Z}
\def\IR{\mathbb R}

\def\shat{^{\mathchoice{}{}%
 {\,\,\smash{\hbox{\lower4pt\hbox{$\widehat{\null}$}}}}%
 {\,\smash{\hbox{\lower3pt\hbox{$\hat{\null}$}}}}}}

\def\bSigma{{
      \ooalign{
      \smash{\hskip.4pt\raise.4pt\hbox{$\Sigma$}}\vphantom{}\crcr
      \smash{\hskip.7pt\raise.6pt\hbox{$\Sigma$}}\vphantom{}\crcr
      \smash{\hbox{$\Sigma$}}\vphantom{$\Sigma$}}
      \vphantom{\hbox{$\Sigma$}}
      }}
\def\bTheta{{
      \ooalign{
      \smash{\hskip.5pt\raise.5pt\hbox{$\Theta$}}\vphantom{}\crcr
      \smash{\hskip.0pt\raise.1pt\hbox{$\Theta$}}\vphantom{}\crcr
      \smash{\hbox{$\Theta$}}\vphantom{$\Theta$}}
      \vphantom{\hbox{$\Theta$}}
      }}
\def\bDelta{{
      \ooalign{
      \smash{\hskip.4pt\raise.4pt\hbox{$\Delta$}}\vphantom{}\crcr
      \smash{\hskip.7pt\raise.6pt\hbox{$\Delta$}}\vphantom{}\crcr
      \smash{\hbox{$\Delta$}}\vphantom{$\Delta$}}
      \vphantom{\hbox{$\Delta$}}
      }}
\def\bLambda{{
      \ooalign{
      \smash{\hskip.5pt\raise.5pt\hbox{$\Lambda$}}\vphantom{}\crcr
      \smash{\hskip.0pt\raise.1pt\hbox{$\Lambda$}}\vphantom{}\crcr
      \smash{\hbox{$\Lambda$}}\vphantom{$\Lambda$}}
      \vphantom{\hbox{$\Lambda$}}
      }}

\makeatletter

\def\bordermatrix#1{\begingroup \m@th
  \@tempdima 8.75\p@
  \setbox\z@\vbox{%
    \def\cr{\crcr\noalign{\kern2\p@\global\let\cr\endline}}%
    \ialign{$##$\hfil\kern2\p@\kern\@tempdima&\thinspace\hfil$##$\hfil
      &&\quad\hfil$##$\hfil\crcr
      \omit\strut\hfil\crcr\noalign{\kern-\baselineskip}%
      #1\crcr\omit\strut\cr}}%
  \setbox\tw@\vbox{\unvcopy\z@\global\setbox\@ne\lastbox}%
  \setbox\tw@\hbox{\unhbox\@ne\unskip\global\setbox\@ne\lastbox}%
  \setbox\tw@\hbox{$\kern\wd\@ne\kern-\@tempdima\left[\kern-\wd\@ne
    \global\setbox\@ne\vbox{\box\@ne\kern2\p@}%
    \vcenter{\kern-\ht\@ne\unvbox\z@\kern-\baselineskip}\,\right]$}%
  \null\;\vbox{\kern\ht\@ne\box\tw@}\endgroup}
\makeatother

\makeatletter
\def\argmin{\mathop{\operator@font arg\,min}}
\def\argmax{\mathop{\operator@font arg\,max}}
\makeatother

\newcommand{\bea}{\begin{array}}
\newcommand{\ena}{\end{array}}
\newcommand{\beq}{\begin{equation}}
\newcommand{\enq}{\end{equation}}

\newcommand{\beqa}{\begin{eqnarray}}
\newcommand{\enqa}{\end{eqnarray}}

\newcommand{\beqan}{\begin{eqnarray*}}
\newcommand{\enqan}{\end{eqnarray*}}

\newcommand{\AL}{\begin{enumerate}}
\newcommand{\ALE}{\end{enumerate}}

\def\addots{\mathinner{
    \mkern1mu\raise0pt\vbox{\kern7pt\hbox{.}}
    \mkern2mu\raise4pt\hbox{.}
    \mkern2mu\raise7pt\hbox{.}
    \mkern1mu}}

\def\sddots{\mathinner{
    \mkern.8mu\raise7pt\hbox{.}
    \mkern.8mu\raise4pt\hbox{.}
    \mkern.8mu\raise0pt\vbox{\kern7pt\hbox{.}}
    \mkern1mu}}

\def\saddots{\mathinner{
    \mkern.2mu\raise0pt\vbox{\kern7pt\hbox{.}}
    \mkern.2mu\raise4pt\hbox{.}
    \mkern.2mu\raise7pt\hbox{.}
    \mkern1mu}}

\def\sqplus{\mathbin{
	{\ooalign{\hfil\raise.3ex\hbox{\scriptsize
	+}\hfil\crcr\mathhexbox274\crcr\mathhexbox275}}
	}} 
\def\sqminus{\mathbin{
	{\ooalign{\hfil\raise.3ex\hbox{\scriptsize
	--}\hfil\crcr\mathhexbox274\crcr\mathhexbox275}}
	}}

\def\IC{{
   \mathord{
      \hbox to 0em{
	 \hskip-4pt
         \ooalign{
	   \smash{\hskip1.9pt\raise2.6pt\hbox{$\scriptscriptstyle |$}}\crcr
	   \smash{\hbox{\rm\sf C}} }
	 \hidewidth}
      \phantom{\hbox{\rm\sf C}}
} }}
\def\IN{
    {\ooalign{
   \smash{\hskip2.2pt\raise1.5pt\hbox{$\scriptscriptstyle |$}}\vphantom{}\crcr
   \hbox{\sf N}
	}}
	} 
\def\IZ{
    {\ooalign{
   \smash{\hskip1.9pt\raise0pt\hbox{$\sf Z$}}\vphantom{}\crcr
   \hbox{\sf Z}
	}}
	} 
\def\IR{
    {\ooalign{
   \smash{\hskip2.2pt\raise1.5pt\hbox{$\scriptscriptstyle |$}}\vphantom{}\crcr
   \smash{\hskip2.2pt\raise3.3pt\hbox{$\scriptscriptstyle |$}}\vphantom{}\crcr
   \hbox{\sf R}
	}}
	} 

\DeclareMathAlphabet{\mathcmb}{OT1}{cmr}{b}{n}

\def\bSigma{\ensuremath{\mathcmb{\Sigma}}}
\def\bLambda{\ensuremath{\mathcmb{\Lambda}}}

\def\bTheta{\ensuremath{\mathcmb{\Theta}}}

\newcommand{\SI}{\begin{indlist}}
\newcommand{\EI}{\end{indlist}}

\newcommand{\DL}{\begin{dashlist}}
\newcommand{\DLE}{\end{dashlist}}

\makeatletter
\def\setboxz@h{\setbox\z@\hbox}
\def\wdz@{\wd\z@}
\def\boxz@{\box\z@}
\def\underset#1#2{\binrel@{#2}%
  \binrel@@{\mathop{\kern\z@#2}\limits_{#1}}}
\def\binrel@#1{\begingroup
  \setboxz@h{\thinmuskip0mu
    \medmuskip\m@ne mu\thickmuskip\@ne mu
    \setbox\tw@\hbox{$#1\m@th$}\kern-\wd\tw@
    ${}#1{}\m@th$}%
  \edef\@tempa{\endgroup\let\noexpand\binrel@@
    \ifdim\wdz@<\z@ \mathbin
    \else\ifdim\wdz@>\z@ \mathrel
    \else \relax\fi\fi}%
  \@tempa
}
\let\binrel@@\relax%
\makeatother

\definecolor{BSUb}{RGB}{0,50,160}     
\definecolor{BSUb2}{RGB}{0,30,96}     
\definecolor{BSUb3}{RGB}{0,114,206}   
\definecolor{BSUo}{RGB}{250,70,22}    
\definecolor{BSUo2}{RGB}{255,130,0}   

\definecolor{custom_green}{HTML}{27ae60}
\usepackage[compact]{titlesec}
\titlespacing{\section}{0pt}{*0.8}{*0.5} 
\titlespacing{\subsection}{0pt}{*0.6}{*0.4}

\usepackage{hyperref}
\usepackage{url}
\usepackage{graphicx}
\usepackage{amsmath}
\usepackage{amsfonts}
\usepackage{booktabs}						
\usepackage{multirow}						
\usepackage{duckuments}						
\usepackage{wrapfig}
\usepackage{ulem}
\usepackage{soul}
\usepackage{xcolor}
\usepackage{enumitem}
\usepackage{subcaption}

\title{RhinoNet: Unsupervised learning for phenoscaping and generation of high-dimensional point clouds}

\author{Siddharth Viswanath$^{*}$\textcolor{custom_green}{$^\star$}, Hiren Madhu$^{*}$\textcolor{custom_green}{$^\star$}, Eddie Cavallin$^{*}$\textcolor{custom_green}{$^\star$}, Benjamin Jones$^{*}$, \\
\textbf{Alistair Wilkinson}$^{*\dagger}$, \textbf{Christopher Tape}$^{\dagger}$, \textbf{Michael Perlmutter}$^{\ddagger}$, \\
\textbf{Rex Ying}$^{*}$, \textbf{Ian Adelstein}$^{*}$, \textbf{Smita Krishnaswamy}$^{*}$ \\[4pt]
$^{*}$Yale University, USA \quad
$^{\dagger}$University College London, United Kingdom \quad
$^{\ddagger}$Boise State University, USA \\
\textcolor{custom_green}{$^\star$}Equal contribution \quad
Corresponding author: \texttt{smita.krishnaswamy@yale.edu}
}

\iclrfinalcopy 
\begin{document}

\maketitle

\begin{abstract}
We present RhinoNet, a simplicial complex-based point cloud neural network, that creates a geometry-preserving embedding of high-dimensional point clouds. RhinoNet takes collections of datasets, also called {\em data cohorts}, and processes each dataset (point cloud) within the cohort as a simplicial complex. These simplicial complexes have datapoints as vertices and edges learned by Vietoris-Rips filtration. Using simplicial wavelets and MLP layers, RhinoNet computes a vector representation of each dataset in a latent space produced by an autoencoder. The latent space is optimized such that an optimal transport distance between pairs of point clouds is preserved between the latent representations. We apply RhinoNet to single-cell datasets collected on patients---a common source of such high-dimensional data. RhinoNet creates a patient level representation, known in biological contexts as a phenoscape. We establish that RhinoNet generates organized phenoscapes, whereas other methods such as static maps do not. Next we show that we can generate entire point clouds by sampling and decoding this phenoscape. We show two modes of generation via 1) latent flow matching and 2) geodesic interpolation in latent space, followed by a decoding back into the point cloud space. RhinoNet achieves state-of-the-art performance on single-cell and spatial data cohorts relative to tested baselines. Overall, RhinoNet lifts the concepts of representation learning and generative modeling to the point cloud or population level and defines a neural model for phenoscaping. 
\end{abstract}

\section{Introduction}\label{sec:intro}
With the advent of high-throughput, high-dimensional data, particularly in biology, a new class of problems has emerged for representation learning. Instead of the classic problem of learning representations of a set of points in a dataset, these problems involve learning representations of entire collections of point clouds, $\mathcal{C} = \{\mathcal{X}^{(1)}, \dots, \mathcal{X}^{(M)}\}$, in which each $\mathcal{X}^{(m)} = \{\mathbf{x}_i\}_{i=1}^{n_m} \subseteq \mathbb{R}^d$ is an unordered dataset of samples from a distribution. Situations where this arises includes single-cell data, or clinical data on cohorts of patients representing disease or conditions. To learn an unsupervised representation in this scenario, an encoder has to map an entire point cloud into a single latent vector, and the organization of the latent space has to reflect population level relationships between datasets, e.g., optimal transport distances. In this paper, we address this problem by proposing a representation learning and generative model called \textbf{RhinoNet}.

\begin{wrapfigure}{r}{0.5\textwidth}
    \centering
    \includegraphics[width=0.5\textwidth]{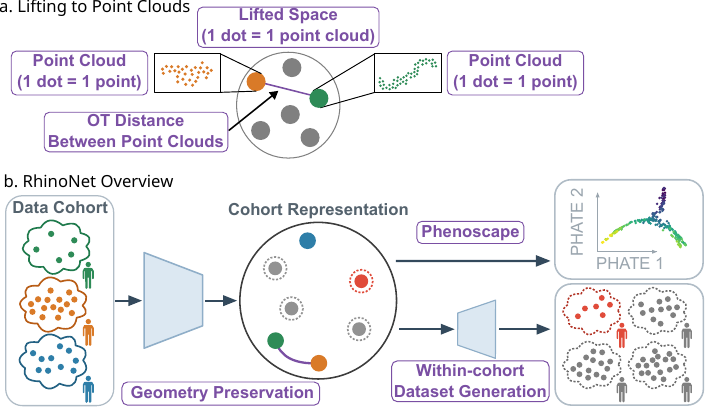}
    \caption{RhinoNet}
    \label{fig:overview}
\end{wrapfigure}

\textbf{RhinoNet} is a simplicial complex-based point cloud autoencoder that lifts representation learning and generation from points to point clouds (Figure~\ref{fig:overview}). We encode point clouds as simplicial complexes, similar to the recently developed HiPoNet framework~\citep{viswanath2025hiponet}. However, in RhinoNet, we are interested in preserving geometric relationships between different point clouds as quantified by  optimal transport distances, analogous to geometric autoencoders for datapoints~\citep{pmlr-v258-sun25c}. RhinoNet begins by taking wavelet transforms of node signals over the simplicial complex utilizing a geometric Hodge Laplacian and an associated diffusion operator in order to preserve more of the signal geometry. It learns its final embedding by further transforming the wavelet representation using an MLP that is penalized with a reconstruction penalty, along with a geodesic Sinkhorn penalty. This construction effectively creates geometry-preserving embeddings that are referred to as {\em phenoscapes}, where each point encodes to an entire dataset and distances between points are interpretable as optimal transport distances. 


Further, RhinoNet is a generative  model with two modes of generation. It can use latent flow matching to generate new point clouds from noise, and it can use geodesic interpolation in latent space to generate new point clouds which are thought of as lying in between two existing point clouds.
%
For example, in a cancer cohort, the first method would generate new point clouds from a similar distribution and the latter would generate new point clouds in between two existing states.

We apply RhinoNet to the problem of phenoscaping cancer patients from single-cell proteomic and transcriptomic datasets collected on each patient. Our results show that RhinoNet most faithfully preserves optimal transport geometry between these patient datasets, and generates accurate point clouds on withheld points. Historically, most generative models for single-cell data do not generate whole point clouds from a latent space, but rather generate new points within a point cloud. Examples include scGen \citep{scgen}, CPA \citep{cpa}, and CellOT \citep{cellot}, so we cannot compare to them. We compare RhinoNet to methods of phenoscaping and point cloud generation. A few methods can both phenoscape and generate whole point clouds. Meta Flow Matching \citep{metaflowmatching} generates a population from a matched control population, and Wasserstein Wormhole both embeds and decodes point clouds. We discuss additional related works in Appendix~\ref{sec:rel-works}.

Our main contributions include: 
\begin{itemize}[leftmargin=*,left=0pt..1em,topsep=0pt,itemsep=-1pt]
    \item \textbf{Geometry-preserving point cloud autoencoder:} We model each point cloud as a simplicial complex and embed it with wavelets built from a geometric Hodge Laplacian, which enables signal processing on the complex in a manner that captures the  underlying geometry.
    \item \textbf{Geodesic Sinkhorn latent regularization:} A novel regularization that jointly enforces a geodesic sinkhorn distance called PhenoGS between point clouds in the latent space.
    \item \textbf{Generative model for point clouds:} We generate whole point clouds by sampling in the latent space. Because latent distances track optimal transport distances, interpolating or extrapolating between the embeddings of observed point clouds and decoding the result yields point clouds for unobserved states of a high-dimensional process.
\end{itemize}
\section{Preliminaries}\label{sec:prelim}
\subsection{Simplicial Complexes and Simplicial Scattering}\label{subsec:simplicial}

\noindent\textbf{Simplicial complexes.} Intuitively, an (abstract) simplicial complex is a higher-order graph that satisfies certain properties. In addition to edges ($1$-simplices) between two objects, a simplicial complex can use higher-order simplices to describe relationships between $3$ or more objects. Simplicial complexes are often used to model hierarchical, relational, and geometric data.

Formally,  we let $\mathcal{V}=\{v_0,v_1,\dots,v_{N_0}\}$ be a finite set of nodes (also known as vertices). A \emph{$k$-simplex}, sometimes called a $k$-th order simplex, $\sigma_k=\{v_{i_0},v_{i_1},\dots,v_{i_k}\}$ is a subset of $\mathcal{V}$ with $k+1$ elements. We interpret $0$-, $1$-, and $2$-simplices as nodes, edges, and triangles. A \emph{simplicial complex} $\mathcal{S}$ is a finite collection of simplices closed under inclusion, meaning every non-empty subset of $\sigma_k \in \mathcal{S}$ is also in $\mathcal{S}$ \citep{schaub2020random, topological-simplex-sp}. We will let  $N_k$ denote the number of $k$-simplices in $\mathcal{S}$ and define the \emph{order} $K$ of the complex $\mathcal{S}$ to be the order of its largest simplex. We will assume that we are given  $D_k$ features for each $k$-simplex and will organize these features into a feature matrix $\mX_k \in \mathbb{R}^{N_k \times D_k}$. We let $\mX = \{\mX_0, \dots, \mX_K\}$ denote the set of all feature matrices. We note that when $K = 1$,  $\mathcal{S}$ is equivalent to an unweighted graph with node features and edge features stored in the matrices $\mathbf{X}_0$ and $\mathbf{X}_1$, respectively.

\noindent\textbf{Boundary matrices.} Connections between simplices of order $k$ and $k-1$ are stored in the boundary matrix $\mB_k \in \mathbb{R}^{N_{k-1} \times N_k}$, whose $(i,j)$-th entry is given by $(\mathbf{B}_k)_{i,j}=1$ if the $i$-th $(k-1)$-simplex is a subset of the $j$-th $k$-simplex and $(\mathbf{B}_k)_{i,j}=0$ otherwise. We denote the set of all the boundary matrices by $\mB = \{\mB_1, \dots, \mB_K\}$.  The \emph{combinatorial Hodge Laplacian} is defined by $\Delta_k=\mathbf{L}_k^\ell+\mathbf{L}_k^u$, where $\mathbf{L}_k^\ell$ and $\mathbf{L}_k^u$ are the lower and upper Laplacians  $\mathbf{L}_k^\ell=\mathbf B_k^\top \mathbf B_k$ and $\mathbf{L}_k^u=\mathbf B_{k+1}\mathbf B_{k+1}^\top$. By convention, $\mathbf{L}_0^\ell=\mathbf{L}_K^u=0$. We observe that in the case $k=0$,  the combinatorial Hodge Laplacian $\Delta_0$ is merely the standard combinatorial graph Laplacian $\mathbf{L}=\mathbf{D}-\mathbf{A}$.

\noindent\textbf{Simplicial random walks and scattering transforms.}
We will extract multiscale information by letting the node features $\mathbf{X}$ diffuse (i.e., propagate) over $\mathcal{S}$. In each step of propagation, the features of a simplex $\sigma_k$ are updated using the features of its neighbors. Notably, on a simplicial complex, each 
 $k$-simplex  $\sigma_k$ can have four types of neighbors: its \emph{boundary adjacent} neighbors $\mathcal{N}_{\mathcal{B}}(\sigma_k)$ are the $(k-1)$-simplices obtained by removing one node from $\sigma_k$; its \emph{coboundary adjacent} neighbors $\mathcal{N}_{\mathcal{C}}(\sigma_k)$ are the $(k+1)$-simplices obtained by adding one node to $\sigma_k$. Its \emph{lower adjacent} neighbors $\mathcal{N}_{\mathcal{L}}(\sigma_k)$ are the $k$-simplices that $\sigma'_k\in\mathcal{S}$ that share a common face, $|\sigma_k\cap\sigma'_k|=k$. Its \emph{upper adjacent} neighbors $\mathcal{N}_{\mathcal{U}}(\sigma_k)$ are the $k$-simplices $\sigma_k'\in\mathcal{S}$ that are both faces of a common $k+1$ simplex, $\sigma_k\cup\sigma_k'\in \mathcal{S}$  and $|\sigma_k\cup\sigma_k'|=|\sigma_k|+1=k+2$. 

Thus, similar to \citet{simplicial-scattering}, we will  introduce  five matrices, one corresponding to each of the four types of neighbors $\mathcal{C}, \mathcal{B},\mathcal{L},$ and $\mathcal{U}$, and a fifth corresponding to self-loops:
\begin{equation}
\begin{gathered}
    \mP^{\mathcal{C}}_{k+1} = \tfrac{1}{5}(\mathbf{D}^{\mathcal{B}}_{k+1})^{-1}
    \mB_{k+1}, \qquad
    \mP^{\mathcal{B}}_{k} = \tfrac{1}{5}\mB_k^{\top}
    (\mathbf{D}^{\mathcal{B}}_{k})^{-1}, \qquad
    \mP^{\cI}_{k} = \tfrac{1}{5}\mathbf{I}_{N_k}, 
    \\[2pt]
    \mP^{\mathcal{L}}_{k} = \tfrac{1}{5}\mB_k^{\top}\mB_k
    (\mathbf{D}^{\mathcal{L}}_{k})^{-1}, \qquad \text{and}\qquad
    \mP^{\mathcal{U}}_{k} =\tfrac{1}{5}\mB_{k+1}\mB_{k+1}^{\top}
    (\mathbf{D}^{\mathcal{U}}_{k})^{-1},
\end{gathered}
\end{equation}
where $\mathbf{D}^{\mathcal{B}}_k = \mathrm{diag}(\mB_k\mathbf{1})$, $\mathbf{D}^{\mathcal{L}}_k = \mathrm{diag}(\mB_k^{\top}\mB_k\mathbf{1})$, and $\mathbf{D}^{\mathcal{U}}_k = \mathrm{diag}(\mB_{k+1}\mB_{k+1}^{\top} \mathbf{1})$ are degree matrices corresponding to different types of neighbors. We note that the normalizing constant $\frac{1}{5}$ is meant to highlight the connection between these matrices and the transition probabilities of a lazy random walker on $\mathcal{S}$. 
We define the diffused node features recursively by first computing
\begin{equation}
\begin{gathered}
    \mX^{(j)}_{\mathcal{B},k} = \mP^{\mathcal{B}}_{k}\mX^{(j-1)}_{k-1}, \qquad
    \mX^{(j)}_{\mathcal{C},k} = \mP^{\mathcal{C}}_{k+1}\mX^{(j-1)}_{k+1},\qquad     \mX^{(j)}_{\cI,k} = \mP^{\cI}_{k}\mX^{(j-1)}_{k}, \\[2pt]
    \mX^{(j)}_{\mathcal{L},k} = \mP^{\mathcal{L}}_{k}\mX^{(j-1)}_{k}, \qquad \text{and} \qquad
    \mX^{(j)}_{\mathcal{U},k} = \mP^{\mathcal{U}}_{k}\mX^{(j-1)}_{k},
\end{gathered}
\end{equation}
and then using a dimension-preserving aggregation:
\begin{equation}
    \mX^{(j)}_{k} = \textsc{AGGR1}\left(\mX^{(j)}_{\cI,k},\,
    \mX^{(j)}_{\mathcal{B},k},\, \mX^{(j)}_{\mathcal{C},k},\, \mX^{(j)}_{\mathcal{L},k},\,
    \mX^{(j)}_{\mathcal{U},k}\right),
\end{equation}
where $\mX^{(0)}_{k} = \mX_k$ and, in our implementation, we take $\textsc{AGGR1}$ to be a simple summation.

These features $\mX^{(j)}_{k}$ can be interpreted as local averages of the original node features at scale $j$ (or as low-pass filterings from the perspective of graph signal processing (GSP), see e.g., \citet{ortega2018graph}). To track changes across scales, we consider wavelets of the form $\psi^{k,\mathcal{N}}_j \mX = \mX^{(j-1)}_{\mathcal{N},k} - \mX^{(j)}_{\mathcal{N},k}$ for each $\mathcal{N} \in \{\cI,\mathcal{B}, \mathcal{C}, \mathcal{L}, \mathcal{U}\}$ (band-pass filtering from the GSP perspective), as developed in \citet{simplicial-scattering}. We then aggregate the wavelet coefficients from different neighbor types to compute simplicial coefficients at scale $j$ as $|\Psi^k_j\mX|=|\textsc{AGGR1}(\psi^{k,\cS\cL}_j, \psi^{k,\mathcal{B}}_j,\psi^{k,\mathcal{C}}_j, \psi^{k,\mathcal{L}}_j, \psi^{k,\mathcal{U}}_j)|$  by applying the modulus (absolute value). We denote the set of embeddings of all orders $|\Psi_j\mX|=\{|\Psi^0_j\mX|, |\Psi^1_j\mX|, \dots, |\Psi^{K}_j\mX|\}$, which has the same form as $\mX$. Applying the same wavelet operators on $|\Psi_j\mX|$ instead of $\mX$ computes the deep simplicial scattering transform $\lvert \Psi_{j'} \lvert \Psi_j \mathbf{X} \rvert \rvert = \left\{ \lvert \Psi_{j'}^{0} \lvert \Psi_j \mathbf{X} \rvert \rvert, \ldots, \lvert \Psi_{j'}^{K} \lvert \Psi_j \mathbf{X} \rvert \rvert \right\}$ where $1\leq j < j'\leq J$. Finally, these coefficients can be combined to compute the embedding of $k$-simplices $S_k(\mX,\mB) = \textsc{COMBINE}(\mX_k, |\Psi^k_j\mX|, \lvert \Psi_{j'} \lvert \Psi_j \mathbf{X} \rvert \rvert, \dots)$ where $\textsc{COMBINE}$ is the concatenation function and the set of embeddings of all orders can be denoted as $S(\mX,\mB)=\{S_k(\mX,\mB), \dots, S_K(\mX,\mB) \}$.
 
 Importantly, we observe that every operator here is assembled directly from the matrices $\mB_k$ whose entries are all either $0$ or $1$. This means that these simplicial scattering coefficients are only able to capture information from the incidence structure, and are unable to capture other types of geometric information. Our methods introduced in Section~\ref{subsec:geometric} will address this limitation allowing us to obtain a more powerful representation of the underlying data geometry.

\noindent\textbf{Simplicial complexes derived from point clouds.} Our data consists of point clouds, each of the form $\mathcal{X} = \{\mathbf{x}_1, \dots, \mathbf{x}_n\} \subseteq \mathbb{R}^d$. We use Vietoris-Rips filtration to simplicial complexes from these point clouds using the. As per~\citet{viswanath2025hiponet}, we first learn  weight vectors $\alpha_{\nu} \in \mathbb{R}^d$, $1\leq \nu \leq V$, and rescale the clouds to give different \emph{views} $\tilde{\mathcal{X}}^{(m)}_{\nu} = \{\alpha_{\nu} \odot \mathbf{x}_i\}_{i=1}^n$ that emphasize different subsets of features, e.g., different cellular processes in single-cell data. Then, we build  a Vietoris--Rips complex $\mathcal{S}^{\nu}$ on each view through a differentiable filtration, yielding boundary matrices $\tilde{\mathbf{B}}^{\nu}$ and simplex features $\tilde{\mathcal{X}}^{(m)}_{\nu}$. Finally, we apply the simplicial scattering transform discussed above to each complex and aggregate to obtain $\Phi = \textsc{AGGR2}\{S(\tilde{\mathbf{X}}^{(1)}, \tilde{\mathbf{B}}^{(1)}), \dots, S(\tilde{\mathbf{X}}^{(V)}, \tilde{\mathbf{B}}^{(V)})\}$, where $\textsc{AGGR2}$ is concatenation across views.

\subsection{Smooth Hodge Laplacian}\label{sec:smoothHodge}

In the smooth setting, 
the Hodge Laplacian depends on the Riemannian metric $g$. This motivates our method, which we will introduce below, where we use  mass matrices $\mathbf{M}_k$ to approximate this metric. We will then use the $\mathbf{M}_k$ to compute a geometric (discrete) Hodge Laplacian.

Formally, for an oriented $n$-dimensional Riemannian manifold $(\mathcal{M}^n,g)$, we denote the (metric-free) exterior derivative on the space of smooth differential $k$-forms by $d:\Omega^{k-1}\to\Omega^k$, and denote its $L^2$ adjoint, 
the co-differential, $\delta:\Omega^k\to\Omega^{k-1}$. 
 The \emph{smooth Hodge Laplacian} is then defined by  $\Delta_k^{\mathrm{smooth}}=d\delta+\delta d$ (note the similarity with the combinatorial Hodge Laplacian for simplicial complexes introduced in Section \ref{subsec:simplicial}, $\Delta_k = \mathbf B_k^\top \mathbf B_k+\mathbf B_{k+1}\mathbf B_{k+1}^\top$).

By the definition of  the adjoint, $\delta$  satisfies $\langle d\eta,\omega\rangle_{L^2(\Omega^k)}=\langle\eta,\delta\omega\rangle_{L^2(\Omega^{k-1})}$ for all $\eta\in\Omega^{k-1},\omega\in\Omega^{k}$, where,  the $L^2$ inner products (and hence $\delta$), depends on the metric $g$. In order to compute these inner products, we will use the \emph{Hodge star} map $\star_g:\Omega^k\to\Omega^{n-k}$, which is characterized by $\omega\wedge\star_g\eta=\langle\omega,\eta\rangle_{g}\, d\mathrm{vol}_g${, where $\langle\cdot,\cdot\rangle_g$ denotes the metric-induced pointwise inner product on $k$-forms.} This allows us to  calculate $L^2$ inner products via $\langle\omega,\eta\rangle_{L^2}=\int_{\mathcal{M}}\omega\wedge\star_g\eta$. We also can use $\star_g$ to characterize the co-differential via $\delta=(-1)^{k(n+1)+1}\star_g d\star_g$. 

\subsection{The Geometric Hodge Laplacian}
\label{sec:DEC and FEEC}
As in \citet{viswanath2025hiponet}, our datasets will consist of point clouds of the form  $\mathcal{X} = \{\mathbf{x}_1, \dots, \mathbf{x}_n\} \subseteq \mathbb{R}^d$, and we will build simplicial complexes and subsequently Hodge Laplacians from these point clouds. However, 
 we will use methods based on a \emph{geometric Hodge Laplacian} $\widetilde{\Delta}_k$, taking inspiration from Discrete Exterior Calculus (DEC)~\cite{desbrun2005discrete} and Finite Element Exterior Calculus (FEEC)~\citep{arnold2018finite}. This is in contrast to   \citet{viswanath2025hiponet} which use methods based on  the combinatorial Hodge Laplacian $\Delta_k$ (see Section \ref{subsec:simplicial}). This change is motivated by the observation that $\Delta_k$ does not depend on any metric and therefore does not in general converge to the smooth Hodge Laplacian 
as $n\rightarrow \infty$. 

To construct the discrete geometric Hodge Laplacian, we let $C^k$ denote the 
set of functions defined on the $k$-simplices  in $\mathcal{S}$. We will identify each $\mathbf{x}\in C^k$ with a vector of length $N_k$ and interpret it  as a $k$-cochain, i.e., as the  discrete integrals of a $k$-form over $k$-dimensional (geometric) simplices.  
The 
 coboundary matrix $\mB_k^\top$ can now be interpreted as a map  $\mB_k^\top:C^{k-1}\to C^k$ that plays the role analogous to the differential $d$ in the smooth setting. As the discrete counterpart of the codifferential $\delta$, we will consider the adjoint $\left(\mB_k^\top\right)^*=\mM_{k-1}^{-1}\mB_k \mM_k$ with respect to the requisite inner product    $\langle \mathbf x,\mathbf y\rangle_{\mM_k}=\mathbf x^\top \mM_k \mathbf y$, where $\mathbf{M}_k$ is a symmetric positive definite matrix to be specified below.
We now define the \emph{geometric Hodge Laplacian} by  $$\widetilde\Delta_k=\mB_k^\top(\mB_k^\top)^*+(\mB_{k+1}^\top)^*\mathbf{B}_{k+1}^\top=\mB_k^\top\, \mM_{k-1}^{-1}\, \mB_k\, \mM_k + \mM_k^{-1}\, \mB_{k+1}\, \mM_{k+1}\, \mB_{k+1}^\top$$
 mirroring the definition $\Delta_k^{\mathrm{smooth}}=d\delta+\delta d$  from  Section \ref{sec:smoothHodge}.  We observe that, by construction, $\widetilde{\Delta}_k$ is  self-adjoint with respect to the inner product $\langle\cdot,\cdot\rangle_{\mM_k}$ and that it reduces to the combinatorial Hodge Laplacian if $\mM_k=\mathbf{I}$.

We will choose the matrix $\mathbf{M}_k\in\mathbb{R}^{N_k\times N_k}$ 
to be a diagonal \emph{mass matrix} inspired by DEC with entries corresponding to the volumes of the associated simplices.
Recent work has proven that the DEC Hodge Laplacian with suitable mass matrices converge to the smooth Hodge Laplacian under mesh refinement when the simplicial mesh and underlying manifold satisfy centering or regularity conditions~\citep{zhu2025convergence, guzman2025framework}. These results connect the DEC constructions to well-established FEEC convergence results~\citep{arnold2010finite, arnold2018finite}. Numerical experiments have also shown that the DEC Hodge Laplacian constructions converge to the smooth Hodge Laplacian, even for simplicial meshes that do not satisfy regularity conditions~\citep{mohamed2016discrete,mohamed2018numerical}. These results 
help motivate our
use of a geometric Hodge Laplacian approach to encode metric information.


\section{Methods}\label{sec:method}

\begin{figure}[h]
    \centering
    \includegraphics[width=\textwidth]{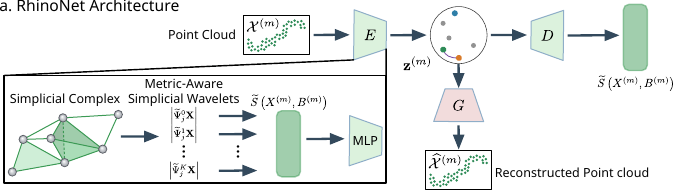}
    \caption{The RhinoNet autoencoder builds a simplicial complex for each point cloud, computes metric-aware simplicial wavelets, and embeds them with an MLP to a latent space. The generation head $G$ is trained to reconstruct point clouds. }
    \label{fig:arch}
\end{figure}

This section describes \textbf{RhinoNet} in two parts: We first introduce an autoencoder that maps each point cloud to a latent vector using a geometric Hodge Laplacian-based simplicial scattering transform. The latent space is regularized by a geodesic Sinkhorn distance between point clouds. We then build on this latent space to generate new point clouds, through latent flow matching and through interpolation. The architecture is depicted in Figure \ref{fig:arch}.

\subsection{Geometric Hodge Laplacian-Based Simplicial Wavelet Autoencoder}\label{subsec:geometric}

Here we describe the \textbf{RhinoNet} autoencoder, whose encoder maps a point cloud $\mathcal{X}^{(m)}$ to a multiscale geometry-aware latent vector $\mathbf{z}^{(m)}$. First, we construct a simplicial complex $\cS^{(m)}$ from the point cloud 
via a Vietoris-Rips filtration. Then, we apply the geometric Hodge Laplacian-based simplicial scattering transform, described below, over this structure to obtain a multiscale representation $\Phi^{(m)}$, which an MLP maps to the latent vector $\mathbf{z}^{(m)}$. The latent space is regularized to enforce a geodesic Sinkhorn distance between point clouds. The decoder then reconstructs $\Phi^{(m)}$ from $\mathbf{z}^{(m)}$.

\textbf{Geometric Hodge Laplacian-based simplicial scattering.} 
We introduce a novel form of the simplicial scattering transform based on the geometric Hodge Laplacian $\widetilde{\Delta}_k$ (see Section \ref{sec:DEC and FEEC}) building on previous work \citep{simplicial-scattering,viswanath2025hiponet} which implicitly used the combinatorial Hodge Laplacian $\Delta_k$. 

In our experiments, we will use simplices of order at most two. Therefore,  we need to specify volumes for $0$-, $1$-, and $2$-simplices in order to define the diagonal mass matrices $\mathbf{M}_0$, $\mathbf{M}_1$, and $\mathbf{M}_2$. We give each $0$-simplex (i.e., vertex) the uniform volume $\frac{1}{N_0}$. We define the volume of a $1$-simplex (edge) between $\mathbf{x}_i$ and $\mathbf{x}_j$  to be its (Euclidean) length $\|\mathbf{x}_i-\mathbf{x}_j\|_2$, and let the volume of each $2$-simplex, $\sigma_\ell^2$, be the area of the triangle it spans, $|\sigma_\ell^2|$. For higher order simplices, Cayley-Menger volumes can be used.  With these choices, we obtain $\mM_0=\frac{1}{N_0}\mathbf{I}$, $\left(\mM_1\right)_{\ell,\ell}=\|\mathbf{x}_i-\mathbf{x}_j\|_2$ when $\sigma_\ell^1$ is the edge between $\mathbf{x}_i$ and $\mathbf{x}_j$, and $\left(\mM_2\right)_{\ell,\ell}=|\sigma_\ell^2|$, when $\sigma_\ell^2$ is a $2$-simplex.

We modify the transition matrices introduced in Section \ref{subsec:simplicial} to incorporate the mass matrices $\mathbf{M}_k$.
This yields metric-aware transition matrices:
\begin{equation}
\begin{gathered}
    \widetilde\mP^{\mathcal{C}}_{k+1} = \tfrac{1}{5}(\mathbf{\widetilde D}^{\mathcal{B}}_{k+1})^{-1}
    \mM_{k-1}^{-1}\mB_{k+1}\mM_k, \qquad
    \widetilde\mP^{\mathcal{B}}_{k} = \tfrac{1}{5}\mB_k^{\top}
    (\mathbf{\widetilde D}^{\mathcal{B}}_{k})^{-1}, \qquad
    \widetilde\mP^{\mathcal{I}}_{k} = \tfrac{1}{5}\mathbf{I}_{N_k}, \\[2pt]
    \widetilde\mP^{\mathcal{L}}_{k} = \tfrac{1}{5}\mB_k^{\top}\mM_{k-1}^{-1}\mB_k\mM_k
    (\mathbf{\widetilde D}^{\mathcal{L}}_{k})^{-1}, \qquad
    \widetilde\mP^{\mathcal{U}}_{k} = \tfrac{1}{5}\mM_k^{-1}\mB_{k+1}\mM_{k+1}\mB_{k+1}^{\top}\left(\mathbf{\widetilde D}^{\mathcal{U}}_{k}\right)^{-1},
\end{gathered}
\end{equation}
with $\mathbf{\widetilde D}^{\mathcal{B}}_{k+1}= \mathrm{diag}(\mM_{k-1}^{-1}\mB_{k+1}\mM_k\mathbf{1})$, $\mathbf{\widetilde D}^{\mathcal{B}}_{k} = \mathrm{diag}(\mB_k^{\top}\mathbf{1})$, $\mathbf{\widetilde D}^{\mathcal{L}}_{k}=\mathrm{diag}(\mB_k^{\top}\mM_{k-1}^{-1}\mB_k\mM_k\mathbf{1})$, and $\mathbf{\widetilde D}^{\mathcal{U}}_{k}=\mathrm{diag}(\mM_k^{-1}\mB_{k+1}\mM_{k+1}\mB_{k+1}^{\top}\mathbf{1})$. We use these transition matrices to compute $\Phi^{(m)}=\widetilde{S}(\mX^{(m)},\mB^{(m)})$, where $\widetilde{S}$ is the wavelet transform defined in Section~\ref{subsec:simplicial} with each $\mP^{\mathcal{N}}_k$ replaced by $\widetilde{\mP}^{\mathcal{N}}_k$.

\textbf{Geodesic Sinkhorn regularization.} The multiscale representation  described above is an initial vector encoding of the entire point cloud. We then use an MLP to further encode $\Phi^{(m)}$ (the vector encoding of the $m$-th point cloud) in a latent space as $\mathbf{z}^{(m)}=\text{MLP}(\Phi^{(m)})=E(\mathcal{X}^{(m)})$. 
In order to produce a structured, low-dimensional representation, we regularize the RhinoNet encoder so that the Euclidean distance between latent points $\mathbf{z}^{(m)}$ and $\mathbf{z}^{(m')}$ matches the PhenoGS~\citep{wilkinson2026phenogsphenoscapescalegeodesicsinkhorn} distance $W$ between the point clouds $\mathcal{X}^{(m)}$ and $\mathcal{X}^{(m')}$, which uses the geodesic Sinkhorn distance. To this effect, we minimize:
\begin{equation*}
    \mathcal{L}_{GS} = \sum_{m,m'}\big(\|\mathbf{z}^{(m)} - \mathbf{z}^{(m')}\|_2 - W(\mathcal{X}^{(m)}, \mathcal{X}^{(m')})\big)^2.
\end{equation*}
We pair the RhinoNet encoder with a decoder $D_\theta$, chosen to be an MLP, that reconstructs the wavelet representation $\Phi$ from the latent $\mathbf{z}^{(m)}$. Our decoder is penalized by the  reconstruction loss in order to  preserve structural information in the latent space: $\mathcal{L}_{rec} = \|D_\theta(\mathbf{z}^{(m)}) - \Phi\|^2_2$.
The encoder and decoder are trained jointly by minimizing $\mathcal{L}_{AE}=\mathcal{L}_{rec}+\lambda\mathcal{L}_{GS}$, where $\lambda$ is a hyperparameter, so that the resulting latent space preserves structural information and the optimal transport geometry between the underlying point clouds. Such an organization of the latent space enables the generation of high-dimensional point clouds, which we describe in the next section. 


\subsection{Data Generation}\label{sec:datagen}
\begin{wrapfigure}{r}{0.5\textwidth}
  \vspace{-12pt}
  \centering
  \includegraphics[width=0.5\textwidth]{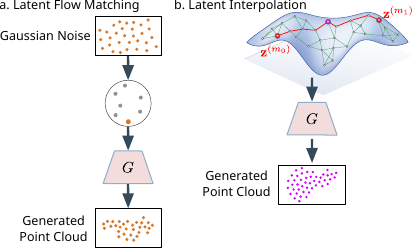}
  \vspace{-16pt}
  \caption{\scriptsize\textbf{Point cloud generation with RhinoNet.} (a) Using latent flow matching from noise. (b) Interpolating in latent space along discrete geodesics built from a $k$-nearest neighbors graph.}
  \label{fig:generation}
  \vspace{-10pt}
\end{wrapfigure}
\noindent\textbf{Generation head.} Here we describe the generation of high-dimensional point clouds from the learned latent space. 
After training the encoder $E$ to map each point cloud $\mathcal{X}^{(m)}$ to a latent $\mathbf{z}^{(m)} = E(\mathcal{X}^{(m)}) \in \mathbb{R}^D$, we consider two modes of generation, as shown in Figure \ref{fig:generation}: sampling new point clouds via latent flow matching, and interpolating between conditions along the latent space. In both cases, a generation head $G_\theta:\mathbb{R}^D \rightarrow \mathbb{R}^d$ maps the latent point $\mathbf{z}^{(m)}$ to a point cloud $\hat{\mathcal{X}}^{(m)} = G_\theta(\mathbf{z}^{(m)})$. We train the generation head by minimizing the Earth mover's distance between the generated point cloud $\hat{\mathcal{X}}^{(m)}$ and the true point cloud $\mathcal{X}^{(m)}$, approximated with entropic optimal transport $\mathcal{W}_\epsilon$ \citep{benamou2015iterative}:
\begin{equation*}
    \mathcal{L}_{gen} = \mathcal{W}_\epsilon(\hat{\mathcal{X}}^{(m)}, \mathcal{X}^{(m)}),
\end{equation*}
where $\epsilon$ is the entropic regularization strength of the Sinkhorn approximation.

\textbf{Latent flow matching.} To generate new point clouds conditioned on a set of labels (e.g., patient, treatment or recurrence), we adopt flow matching~\citep{lipman2023flow}, which learns a velocity field that transports a standard Gaussian source to the distribution of the observed latents  for a given combination of labels and is simulation-free during training. Conditioning the velocity field on the labels allows us to sample latents for combinations that were never directly observed during training. At inference time, we integrate the learned velocity starting from Gaussian noise to obtain a sampled latent for a target combination of labels, which we then decode into a point cloud through the generation head $G_\theta$. 

\textbf{Interpolation.} We now consider the setting where want to be able to move between two points in the latent space, which may, for example, correspond to two different conditions in a patient. 
We consider a path between the two latent points and interpret points along the path as intermediate states 
so that the latent space forms a continuous phenoscape. Since the true trajectory between two conditions need not be straight in latent space, we construct the path as a discrete geodesic over the graph of observed latent points. Recall that the encoder is trained so that Euclidean distances between latent points approximate optimal transport distances between the corresponding point clouds. The shortest path between two latent points should therefore approximate the shortest path between their corresponding point clouds under optimal transport. We build a symmetrized $k$-nearest neighbor graph over all the $N$ latents $\{\mathbf{z}^{(m)}\}_{m=1}^M$, where each vertex corresponds to an observed point cloud $\mathcal{X}^{(m)}$. The graph distance between two point clouds is then the shortest path between them in this graph computed via Dijsktra's algorithm~\citep{dijkstra1959note}.We select two point clouds $\mathcal{X}^{(m_0)}$ and $\mathcal{X}^{(m_1)}$, drawn from two different conditions to serve as the path endpoints. Let $c(m)$ denote a label of the point cloud $\mathcal{X}^{(m)}$, so that $c(m_0) \neq c(m_1)$. The discrete geodesic between $\mathbf z^{(m_0)}$ and $\mathbf z^{(m_1)}$ then gives a sequence of point clouds connecting the two conditions, with interior point clouds serving as intermediate states. To evaluate whether these intermediate point clouds capture meaningful structure, we hold them out and evaluate whether the decoder can reconstruct each one from its corresponding latent.

RhinoNet is trained in two stages. We first train the autoencoder by minimizing $\mathcal{L}_{AE}$ to obtain a latent space organized by geodesic Sinkhorn distance between point clouds. We then train the generation head $G_{\theta}$ and the components used for latent flow matching and interpolation on top of this latent space. Computational complexity of the method is discussed in Appendix~\ref{app:complexity}. 
\vspace{-1em}
\section{Results}\label{sec:res}

In this section, we describe the datasets used to train RhinoNet, the tasks we evaluate on, and the baselines we compare against, followed by results and ablation studies. Because our latent space is organized geometrically, we evaluate it using metrics that directly compare point cloud geometry. For the interpolation task, where a ground-truth intermediate point cloud is available, we measure the Earth Mover's Distance (EMD) between the generated point cloud and this ground truth. For the latent flow matching task, where new point clouds are generated from noise, we instead compare the distribution of generated point clouds to the distribution of real point clouds using 1-Wasserstein distance ($\mathcal{W}_1)$ a.k.a, the Earth mover's distance, 2-Wasserstein distance ($\mathcal{W}_2$), and Maximum Mean Discrepancy (MMD). Further details on the experimental setup are provided in Appendix~\ref{app:expsetup}.

\noindent\textbf{Data cohorts.} We evaluate RhinoNet on a synthetic Swiss-roll dataset and on two real-world biological data cohorts. First, we use a 10-dimensional synthetic Swiss-roll with a known ground truth manifold structure as a controlled setting for evaluating neighborhood preservation in the learned latent space. We then evaluate RhinoNet on two real-world biological data cohorts: the first is a set of patient-derived organoids (PDOs)~\citep{trellis-pdo}, consisting of organoid cultures grown from colorectal cancer patients, with each culture represented as a point cloud of cells characterized by a panel of 44 protein markers; the second is CODEX-based spatial transcriptomics data~\citep{space-gm}, spanning multiple patient cohorts including Charville and UPMC with head-and-neck or colorectal cancer, where each cell's protein expression is measured across 40 markers along with its spatial position in the tissue. Further details on the data cohorts, including cohort sizes, marker panels, and preprocessing are provided in Appendix~\ref{app:data}.

\noindent\textbf{Baselines.} We evaluate RhinoNet against a set of baselines for each task. For the phenoscape evaluation, we compare against DiffusionEMD~\citep{pmlr-v139-tong21a}, PointNet++~\citep{pointnet++}, PointTransformer~\citep{pointtransformer}, and Meta Flow Matching~\citep{metaflowmatching}. For the generation of single-cell data using latent flow matching evaluation, we compare against Point-Voxel Diffusion (PVD)~\citep{zhou20213dshapegenerationcompletion}, PointFlow~\citep{pointflow}, and LION~\citep{zeng2022lion}. For the interpolation task, we compare against Meta Flow Matching~\citep{metaflowmatching}.


\begin{wrapfigure}{r}{0.5\textwidth}
  \vspace{-10pt}
  \centering
  \includegraphics[width=0.5\textwidth]{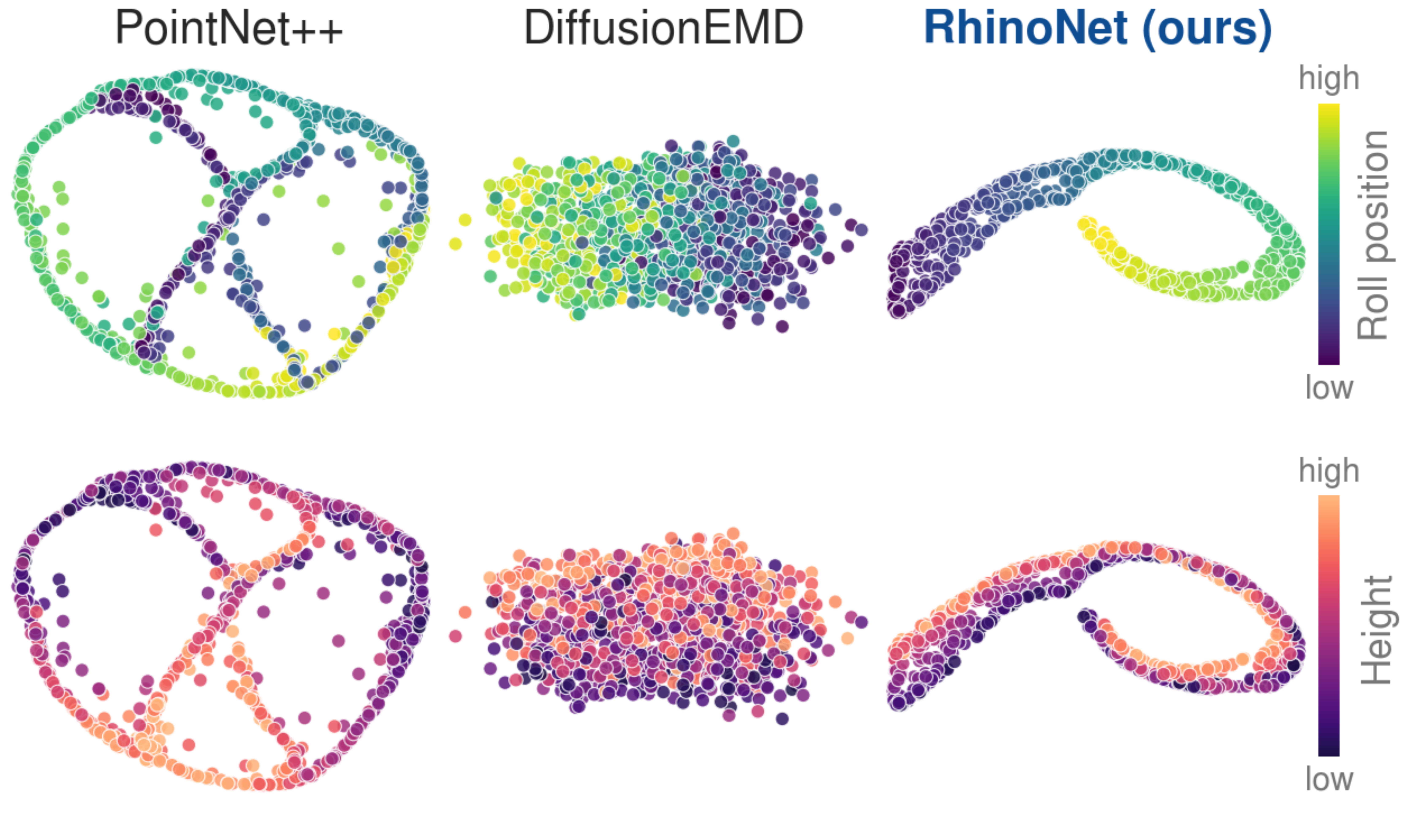}
  \vspace{-14pt}
  \caption{\scriptsize\textbf{RhinoNet preserves 10D Swiss roll geometry.} 2D PHATE of each method's latent space, one dot per point cloud, colored by position along the roll (top) and height (bottom).}
  \label{fig:swiss_roll}
  \vspace{-8pt}
\end{wrapfigure}
\noindent\textbf{RhinoNet can preserve the manifold geometry.} We first evaluate how well the geometry learned by RhinoNet captures the structure of a manifold, comparing against DiffusionEMD, PointNet++, and PointTransformer. Each method maps a point cloud to a single latent vector, which we reduce to two dimensions with PHATE~\citep{moon2019visualizing}, so that every dot in Figure~\ref{fig:swiss_roll} is one entire point cloud, colored by a known property: the generating sheet coordinates for the Swiss roll, and sample metadata or mean marker expression for the biological cohorts. The Swiss roll cohort consists of 1,000 point clouds whose centers lie on a two-dimensional sheet rolled up in three dimensions and embedded in $\mathbb{R}^{10}$, with ground-truth distance measured along the unrolled sheet. RhinoNet recovers a single band ordered by position along the roll and retains height as a second axis. PointNet++ folds the roll onto itself, placing point clouds from distant parts of the sheet side by side, and DiffusionEMD keeps only a coarse ordering with no local structure. Neither PointNet++ nor DiffusionEMD preserves height.

\begin{figure}[t]
  \centering
  \includegraphics[width=\textwidth]{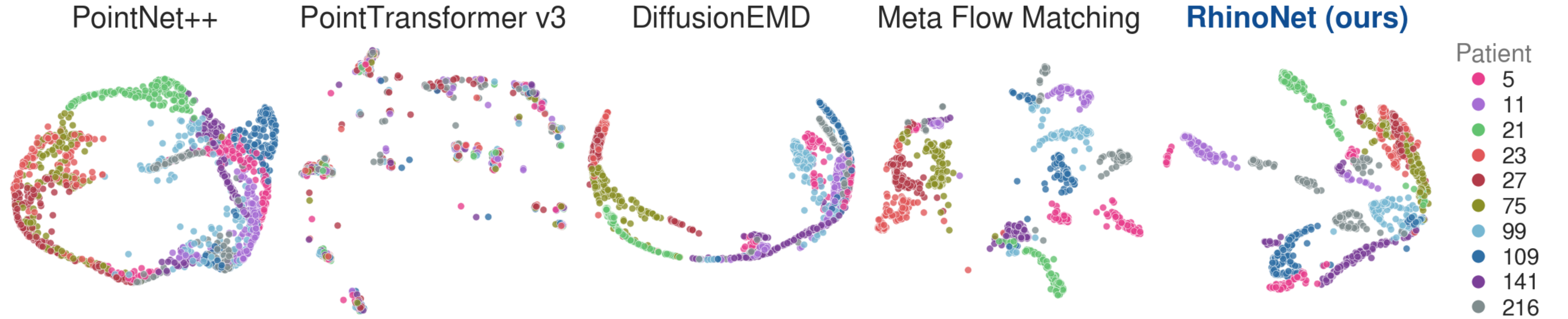}
  \caption{\footnotesize \textbf{RhinoNet organizes patient-derived organoids by patient.} Phenoscapes of 1678 colorectal cancer organoid samples from 10 patients (2M cells, 44 protein markers). Each dot is one dataset, visualized using PHATE of that method's latent space and colored by patient of origin.}
  \label{fig:pdo_phenoscape}
\end{figure}

\noindent\textbf{Phenoscape.} We next construct phenoscapes of the patient-derived organoid PDO cohort \citep{trellis-pdo}. Each sample is a point cloud of roughly two thousand cells measured across patients. Here, we ask whether the phenoscape recovers the dominant source of biological variation: the patient of origin. Figure~\ref{fig:pdo_phenoscape} shows that RhinoNet places most patients on distinct branches of the phenoscape, including held-out patients, which were embedded without being seen during training. PointNet++ arranges samples along a single ring on which several patients overlap. DiffusionEMD orders samples along one curve and only partially separates patients. PointTransformer collapses the cohort into a few isolated clusters that mix patients. While MFM separeted the patients well, it places each patient in a compact cluster, whereas RhinoNet arranges each patient's samples along a continuous branch. This difference likely reflects that Meta Flow Matching trains its point cloud embedding to condition a flow, with no term relating distances between embeddings to distances between point clouds. Together with the Swiss roll results, this shows that the geodesic Sinkhorn regularisation produces a latent space whose large-scale organisation reflects the geometric structure of the cohort. We show a similar phenoscape for Charville in Figure~\ref{fig:charville_phenoscape} and UPMC~\ref{fig:upmc_phenoscape} colored by the gene expression levels showing a gradient in expression level across the phenoscape while other methods fail to encode this kind of structure. 

\noindent\textbf{Generating single-cell datasets using Latent Flow Matching.}
We next assess RhinoNet's ability to generate single-cell point clouds using the latent flow-matching procedure described in
Section~\ref{sec:datagen}. For the PDO data cohort, generation is conditioned on patient identity and treatment, enabling the model to generate a cellular point cloud associated with a specified patient--treatment combination. For the Charville and UPMC cohorts, generation is conditioned on patient identity and observed recurrence status.
As shown in Table~\ref{tab:emd_w2_mmd}, RhinoNet achieves the lowest values for all three metrics on PDO and Charville. Overall, these results indicate that RhinoNet produces point clouds with strong marker-space and distributional agreement across the three cohorts, while PointFlow remains competitive on UPMC.

\begin{table}[h]
\centering
\caption{\footnotesize Comparison of generated point clouds across the PDO, Charville, and UPMC data cohorts using 1-Wasserstein distance, 2-Wasserstein distance, and Maximum Mean Discrepancy (MMD) between the generated and real point cloud distributions. Lower values indicate better generation quality. Bolded score indicates best overall performance, and an underlined score indicates second-best performance.}
\label{tab:emd_w2_mmd}
\resizebox{\linewidth}{!}{
\begin{tabular}{lccccccccc}
\toprule
Model
& \multicolumn{3}{c}{PDO}
& \multicolumn{3}{c}{Charville}
& \multicolumn{3}{c}{UPMC} \\
\cmidrule(lr){2-4} \cmidrule(lr){5-7} \cmidrule(lr){8-10}
& EMD ($\cW_1$) $\downarrow$ & $\cW_2$ $\downarrow$ & MMD $\downarrow$
& EMD ($\cW_1$) $\downarrow$ & $\cW_2$ $\downarrow$ & MMD $\downarrow$
& EMD ($\cW_1$) $\downarrow$ & $\cW_2$ $\downarrow$ & MMD $\downarrow$ \\
\midrule
PVD
& 36.8437 & 119.8368 & 0.249434
& 35.3825 & 50.1669 & 0.286170
& 12.5308 & 12.6588 & 0.263725 \\
PointFlow
& \underline{6.1260} & \underline{6.2154} & \underline{0.024944}
& \underline{4.7665} & \underline{4.9663} & \underline{0.037828}
& \textbf{2.8926} & \underline{3.2072} & \textbf{0.014311} \\
LION
& 66.3286 & 259.9126 & {0.198157}
& 240.1520 & 806.4914 & {0.264518}
& 26.7287 & 35.8295 & 0.371724 \\
\textbf{RhinoNet}
& \textbf{4.2305} & \textbf{4.3241} & \textbf{0.021284}
& \textbf{3.7432} & \textbf{3.9447} & \textbf{0.025637}
& \underline{2.9567} & \textbf{3.1816} & \underline{0.027461} \\
\bottomrule
\end{tabular}
}
\end{table}

\begin{figure}[t]
      \centering
      \includegraphics[width=\textwidth]{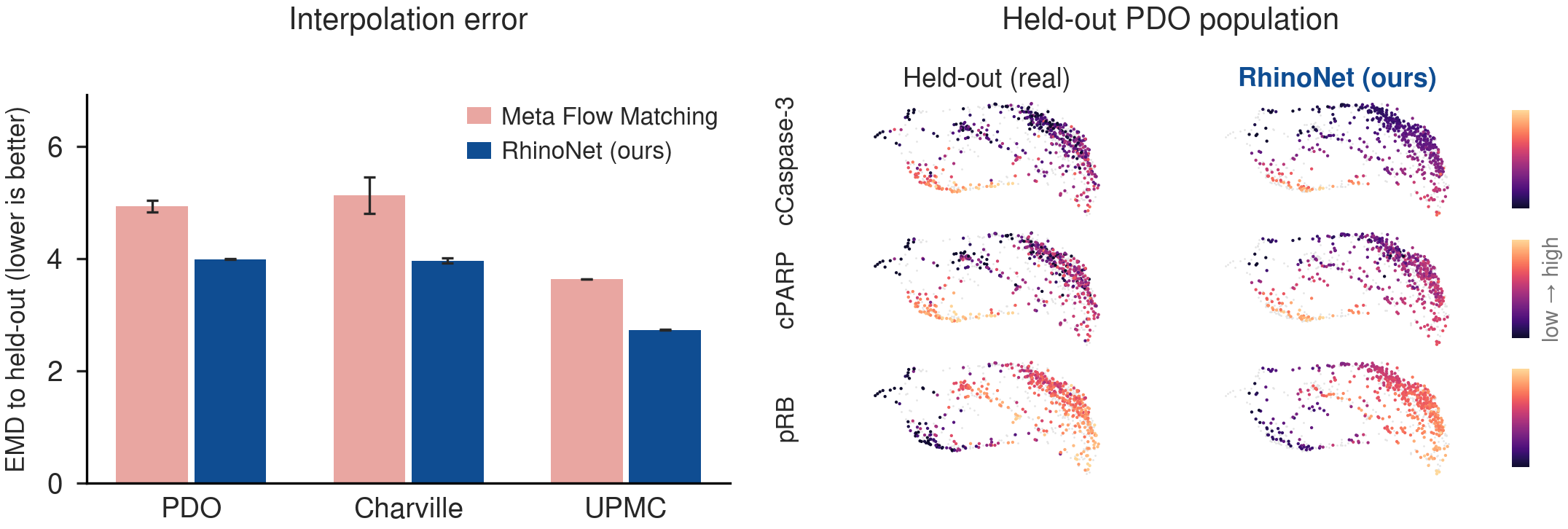}
      \caption{\footnotesize\textbf{Interpolating held-out point clouds.}
      \textbf{Left:} Earth Mover's distance (EMD) between generated and held-out point clouds (lower is better). Bars show the mean $\pm$ s.d.\ over three seeds of the average EMD across each dataset's held-out point clouds. \textbf{Right:} a held-out PDO point cloud from patient 21, lying on the phenoscape path between DMSO and treatment 5-FU. Real cells (left) and cells generated by RhinoNet (right) are embedded jointly with PHATE and colored by cleaved caspase-3, cleaved PARP and phosphorylated RB. Each point in the plot is a cell.}
      \label{fig:interpolation}
\end{figure}

\noindent\textbf{Interpolation.} Finally, we evaluate RhinoNet's ability to interpolate between point clouds in its learned latent space. For each cohort we construct a discrete geodesic between two endpoint point clouds (Section~\ref{sec:datagen}), hold out the intermediate point clouds on the path from decoder training, and reconstruct them from their latent codes. On PDO, the path runs within a single patient from a control (DMSO) culture to a drug-treated culture (treatment F), with three intermediate cultures held out. On Charville and UPMC, the path runs from a non-recurrent to a recurrent sample of the same patient, testing whether the latent space captures a continuous transition associated with disease recurrence; two intermediate samples are held out on Charville and three on UPMC, where the intermediate samples come from three patients. Across all three cohorts, RhinoNet reconstructs held-out intermediate point clouds more accurately than Meta Flow Matching, Figure~\ref{fig:interpolation}, left), reducing EMD by 19\% on PDO (3.98 vs. 4.93), 23\% on Charville (3.96 vs.\ 5.13) and 25\% on UPMC (2.73 vs.\ 3.63). RhinoNet is also more stable across seeds (s.d.\ $\leq 0.05$ vs.\ up to 0.32 for Meta Flow Matching). Beyond matching the overall distribution, the generated cells recover how signaling varies within a held-out organoid point cloud: apoptosis markers (cleaved caspase-3, cleaved PARP) and a proliferation marker (pRB) follow the same gradients across the PHATE embedding as in the real cells (Figure~\ref{fig:interpolation}, right). Figure~\ref{fig:interpolation_paths} (Appendix~\ref{app:interpolation-paths}) shows each path in RhinoNet's phenoscape. In all three cohorts, the geodesic between the two conditions passes through a short run of neighboring point clouds rather than a single long jump. The held-out point clouds are therefore intermediate states between the endpoints, and RhinoNet decodes each one from its latent position on the path.

\begin{wrapfigure}{r}{0.42\textwidth}
      \vspace{-12pt}
      \centering
      \includegraphics[width=0.42\textwidth]{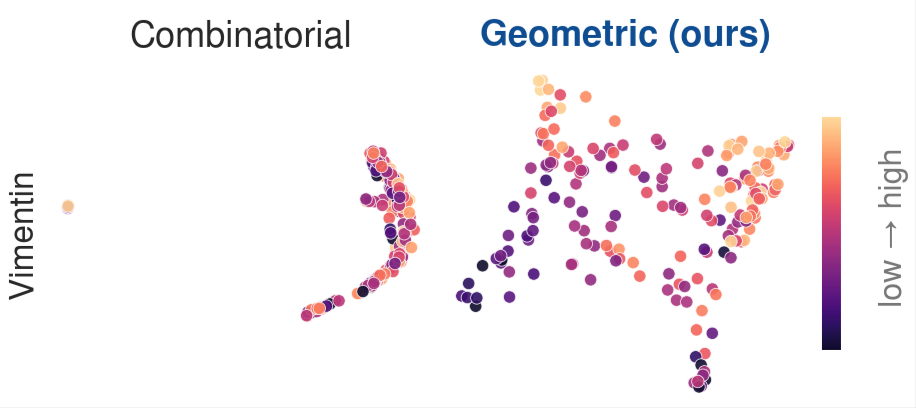}
      \vspace{-14pt}
      \caption{\scriptsize\textbf{Hodge Laplacian ablation.} Charville phenoscapes from RhinoNet with the combinatorial or geometric Hodge Laplacian, colored by mean Vimentin expression per region.}
      \label{fig:laplacian_ablation}
      \vspace{-10pt}
  \end{wrapfigure}

\noindent\textbf{Ablations.} To isolate the effect of the geometric Hodge Laplacian, we retrain RhinoNet on Charville without using the mass matrices, which corresponds to use the combinatorial Laplacian in place of the geometric Laplacian. As we can see in Figure~\ref{fig:laplacian_ablation}, the full version of RhinoNet based on the geometric Laplacian fits the geodesic Sinkhorn distances well, and gives smoother marker gradients across the phenoscape compared to the combinatorial version. We present a detailed description and more markers in Appendix Figure~\ref{fig:laplacian_ablation_full}. Furthermore, increasing the simplex order from $K=1$ (a graph over cells) to $K=2$ (adding triangles) improves the fit to the geodesic Sinkhorn distances on both spatial cohorts. On UPMC it also orders tissue regions along a smoother axis of immune infiltration (Figure~\ref{fig:k_ablation_upmc}), while on Charville both orders recover the immune gradient (Figure~\ref{fig:k_ablation_charville}).

\section{Conclusion}\label{sec:conclusion}

In this work, we present RhinoNet, a novel simplicial complex based point cloud neural network that preserves the geometry of the underlying data. RhinoNet is trained with an autoencoder such that each point in the latent space is one point cloud. RhinoNet's latent space is optimized such that the optimal transport distance between pairs of point clouds is preserved. When applied to single cell data, the resulting patient level phenoscape is organized in a biologically meaningful way. RhinoNet's geometry-aware phenoscape also enables generation of new single cell populations conditioned on patients and treatments. The geodesic interpolation in the latent space decodes back the points accurately. These generated and interpolated point clouds contain meaningful biological signals about disease progression or untried treatments, which are sparsely measured due to cost and time. Hence, reasoning at the level of populations rather than cells makes cohorts of point clouds a data type that can be compared, searched, and generated to progress biological discoveries.

\section*{Acknowledgments}

M.P. acknowledges funding from The National Science Foundation under grant number OIA-2242769. S.K. is funded in part by the NIH (NIGMSR01GM135929, R01GM130847), NSF CAREER award IIS-2047856, NSF IIS-2403317, NSF DMS-2327211 and NSF CISE 2403317. S.K is also funded by the Sloan Fellowship FG-2021-15883, the Novo Nordisk grant GR112933, and the NCI grant OT2CA319802. S.K. and M.P. also acknowledge funding from NSF DMS-2327211. 

\subsection*{AI use statement}
We used large language models (LLMs), including Claude (Anthropic) for writing and coding. Among the tasks that require disclosure, we used LLMs to implement parts of RhinoNet and the baseline adaptations and to give feedback on whether our descriptions of results matched the figures and tables; the interpretations reported in the paper were decided by the authors. We did not use LLMs to generate synthetic datasets, develop theoretical models, formulate or prove mathematical claims, propose hypotheses, design experiments, or clean datasets, and translation and qualitative data analysis are not applicable to this work. Additionally, we used LLMs to draft, edit, and structure parts of the paper, to identify and summarize relevant literature, and to format tables. The authors reviewed and revised all LLM-assisted text, checked every LLM-suggested reference against the original paper, and reviewed and tested all LLM-assisted code and computed values. We take responsibility for the final content of this work, including text, claims, and artifacts produced with the aid of generative AI.

\subsection*{Ethics statement}
This work uses only previously published data and involves no new data collection from human subjects. The patient-derived organoid data \citep{trellis-pdo} and the CODEX data from the Charville and UPMC cohorts \citep{space-gm} are both publicly available data cohorts. RhinoNet generates synthetic single-cell point clouds, including point clouds for conditions that were never measured. These are model outputs, not measurements: they should be labeled as synthetic when shared and should not inform clinical or treatment decisions without experimental validation. Our cohorts include a small number of patients with colorectal or head-and-neck cancer, so patterns learned from them may not generalize to other patient populations, cancer types, or measurement platforms. 

\subsection*{Reproducibility statement}

We have made all code for RhinoNet, the baseline adaptations, and the evaluation available at \url{https://anonymous.4open.science/r/RhinoNet-B4DE/}, and it is also included in the supplementary material. All datasets used in this work are publicly available: the patient-derived organoid data \citep{trellis-pdo} and the CODEX data from the Charville and UPMC cohorts \citep{space-gm} can be obtained from the original studies, and the synthetic Swiss roll can be regenerated with the provided code. Section~\ref{sec:method} describes the architecture, the geometric Hodge Laplacian, and the training objective, and Appendix~\ref{app:data} describes the cohort sizes, marker panels, and preprocessing. Section~\ref{sec:res} describes how each baseline was adapted to high-dimensional marker spaces, and Appendix~\ref{app:expsetup} lists the hyperparameters and training details.

\newpage
\bibliography{iclr2027_conference}
\bibliographystyle{iclr2027_conference}

\newpage
\appendix
\pagenumbering{arabic}
\renewcommand*{\thepage}{A\arabic{page}}
\renewcommand{\thefigure}{A\arabic{figure}}
\renewcommand{\thetable}{A\arabic{table}}
\setcounter{figure}{0}
\setcounter{table}{0}
\setcounter{page}{1}

\section{Related Works}\label{sec:rel-works}
\noindent\textbf{Point Cloud and Set Networks.}
Neural networks on point clouds must map a set of points of any size to a fixed-size representation. Deep Sets \citep{zaheer2017deep} and PointNet \citep{pointnet} apply a shared network to each point and pool the results, and Set Transformer \citep{lee2019set} adds attention between points before pooling. Other encoders compute features over local structure: PointNet++ \citep{pointnet++} groups points into local neighborhoods, DGCNN \citep{dgcnn} applies edge convolutions over a $k$-nearest-neighbor graph recomputed at each layer, and Point Transformer \citep{pointtransformer} applies attention within local neighborhoods. Point cloud autoencoders, such as that of \citet{achlioptas2018learning} and SetVAE~\citep{kim2021setvae}, learn latent spaces by reconstructing each point cloud. PointNet, PointNet++, DGCNN, and Point Transformer were designed for 3D shapes, where the Euclidean distance between points is meaningful. These encoders are typically trained on supervised targets or on per-cloud reconstruction, and neither objective constrains distances between point clouds in the latent space. RhinoNet adds this constraint and targets high-dimensional point clouds.

\noindent\textbf{Simplicial Networks.}
Simplicial complexes extend graphs with higher-order simplices such as triangles, and Hodge Laplacians extend the graph Laplacian to signals on these simplices~\cite{lim2020hodge}. Simplicial neural networks pass messages between incident or adjacent simplices \citep{mpsn, roddenberry2021principled}, and wavelet constructions on graphs~\citep{coifman2006diffusion} have been extended to simplicial complexes~\citep{simplicial-scattering}. HiPoNet \citep{viswanath2025hiponet} applies these tools to point clouds: it builds several simplicial complexes per point cloud from learned feature reweightings, extracts multiscale features with a simplicial wavelet transform, and is trained for supervised tasks. These networks use the combinatorial Hodge Laplacian, which depends only on which simplices are incident to which. RhinoNet's geometric Hodge Laplacian depends on the positions of the points, and RhinoNet is trained without labels.

\noindent\textbf{Optimal Transport Between Point Clouds.}
Optimal transport (OT) compares two point clouds by the cost of moving the points of one onto the other, lifting a ground cost between points to a distance between point clouds, and entropic regularization makes it tractable with the Sinkhorn algorithm~\citep{cuturi2013sinkhorn}. Kernel mean embeddings offer another way to compare distributions and underlie distribution regression \citep{szabo2016learning}. We use OT because its ground cost gives direct control over how distances between points are measured, which we set to a distance along the data manifold. Static maps compute OT distances between all pairs of point clouds in a collection and embed them non-parametrically. PhEMD \citep{chen2020phemd} applies MDS or PHATE \citep{moon2019visualizing} to the pairwise distance matrix, Diffusion EMD \citep{pmlr-v139-tong21a} computes transport distances by diffusion over a graph shared by the collection, and PILOT~\citep{joodaki2023detection} computes OT distances between patients from single-cell data. These embeddings are tied to the collection they were computed on: placing a new point cloud requires new transport problems and a recomputed embedding, and no decoder maps a location back to a point cloud. Deep Wasserstein embeddings \citep{courty2017learning} and Wasserstein Wormhole \citep{haviv2024wasserstein} train an encoder and a decoder so that Euclidean distances between latent points match OT distances, which makes the embedding inductive and invertible, but both use the Euclidean distance between points as the ground cost. RhinoNet likewise solves transport problems only during training, and it replaces the Euclidean ground cost with a geodesic one.

\noindent\textbf{Manifold Geometry and the Ground Cost.}
Under the manifold hypothesis, high-dimensional data lie near a lower-dimensional manifold, and distances along that manifold can differ from ambient Euclidean distances. In high dimensions, Euclidean distances also concentrate under broad conditions, so that nearest and farthest neighbors become hard to tell apart \citep{beyer1999nearest}. Diffusion maps~\citep{coifman2006diffusion} and heat-based geodesics~\cite{crane2013geodesics} approximate manifold distances through diffusion over a graph or mesh, and Heat-Geo \citep{heat-geo} uses heat diffusion to build embeddings of individual points that preserve geodesic distances. Geodesic Sinkhorn \citep{huguet2023geodesic} uses a heat-diffusion geodesic as the ground cost of entropic OT, so that the distance between point clouds follows the manifold. RhinoNet uses geodesic Sinkhorn distances as a training signal for its encoder, so they need not be computed for every pair of point clouds at test time.

\noindent\textbf{Generative Models for Point Clouds.}
Generative models for 3D point clouds learn a distribution over point clouds. \citet{achlioptas2018learning} train GANs in the latent space of a point cloud autoencoder, PointFlow \citep{pointflow} uses continuous normalizing flows over a shape-level latent variable and over points, PVD \citep{zhou20213dshapegenerationcompletion} generates point clouds by iterative denoising, and LION \citep{zeng2022lion} combines a hierarchical variational autoencoder with diffusion models over global and point-structured latent representations. PointGPT \citep{chen2023pointgptautoregressivelygenerativepretraining} pretrains a transformer by autoregressively generating patches of a point cloud. These models were designed for three-dimensional coordinates, and we adapt them to high-dimensional marker spaces as baselines. In single-cell biology, generative models mostly work at the level of points: scGen \citep{scgen}, CPA \citep{cpa}, and CellOT \citep{cellot} predict how individual cells respond to a perturbation, and Meta Flow Matching \citep{metaflowmatching} generates a population from a matched source population.  RhinoNet generates whole point clouds from any location in a geometry-preserving latent space, including locations between observed point clouds.

\section{High Dimensional Point Cloud datasets}\label{app:data}

\noindent\textbf{PDO Data.} The PDO dataset originates from a colorectal cancer drug-screening study by \cite{trellis-pdo}, which profiled 10 patient-derived organoid (PDO cultures using CyTOF under two culture conditions: PDO monoculture and co-culture with cancer-associated fibroblasts (CAFs). To enable high-throughput multiplexing across more than 2,500 3D cultures, the study employed TOBis (thiol-reactive organoid barcoding in situ), measuring a 44-antibody panel spanning PTM signaling, DNA damage response, apoptosis markers, and cell-state and cell-type identifiers. The drug screen covered three standard-of-care chemotherapies, SN-38 (S), 5-fluorouracil (5-FU, F), and oxaliplatin (O), alongside three targeted agents: cetuximab (C), berzosertib/VX-970 (V), and LGK-974 (L) applied as both single agents and combination treatments. For preprocessing, we used the per-cell marker profiles provided by the source study, organizing cells by patient identity, culture condition, and treatment.

\noindent\textbf{Charville Data.} The Stanford-CRC dataset comprises 292 CODEX multiplexed immunofluorescence samples from 161 colorectal cancer patients. They were collected across 4 batches and contain 632,280 segmented and classified cells. Each sample represents a tumor microarray core stained with a 40-plex antibody panel, with single-cell protein expression quantified across all markers and cells assigned to types via unsupervised clustering and k-nearest-neighbour label transfer. Samples are annotated with three clinical outcomes: primary outcome (alive after 60 months: 122 patients; recurred or dead within 60 months: 74), binary recurrence status (recurred: 57; not recurred: 129), and time-to-recurrence interval.

\noindent\textbf{UPMC Data.} The UPMC-HNC dataset covers 81 head and neck cancer patients across 308 tumor microarray samples with 7 batches. It contains over 2 million cells segmented and classified from 40-plex CODEX imaging. Cell identities were established by unsupervised clustering of per-cell protein expression profiles followed by k-nearest-neighbor label transfer across the full dataset. Clinical annotations span four axes: primary tumor outcome (NED: 197; non-NED: 111), overall survival length, disease recurrence (recurred: 31; not recurred: 253), and HPV infection status (positive: 158; negative: 150).

\section{Other Sources of High-Dimensional Data Cohorts}\label{app:other-cohorts}
RhinoNet applies to cohorts in which each experimental unit contributes a point cloud: an unordered set of observations, possibly of different sizes, measured in a feature space shared across the cohort. The questions it addresses concern relationships between units. Suitable cohorts also carry labels for each unit, such as a patient, treatment, site, or time, which can condition generation or colour a phenoscape. The cohorts we evaluate in Section~\ref{sec:res} contain between 186 and 1,678 point clouds, and cohorts with the same structure arise in other fields.
In neuroimaging~\citep{di2014autism}, a multi-site resting-state fMRI study could treat each scanning session as a point cloud whose points are time frames, each described by the signal in several hundred brain regions. In a phenoscape of such a cohort, sessions would lie near one another when their distributions of brain states are similar, and acquisition site and diagnostic group could be examined as separate sources of variation in the latent space.
In ecology~\cite{keller2008continental}, a network of long-term monitoring sites that records many organisms at each sampling visit, each organism annotated with environmental and spectral measurements, would yield one point cloud per site and season. Decoding latent points between consecutive seasons of the same site could then generate that site's community at times that were not sampled.
In finance~\cite{chen2022open}, the monthly cross-section of listed firms, with each firm described by a few hundred accounting and market characteristics, forms one point cloud per month, so a panel spanning several decades gives several hundred point clouds. Months with similar distributions of firm characteristics would lie near one another in the phenoscape, and latent flow matching could sample synthetic cross-sections conditioned on labels such as a recession indicator.
We have not evaluated RhinoNet in any of these settings.

\section{Experimental Setup}
\label{app:expsetup}
The parameters of RhinoNet are optimized using the AdamW optimizer with a constant learning rate of \(10^{-3}\) and weight decay of \(3\times10^{-3}\). We train the PDO model for at most 120 epochs and the Charville and UPMC models for 100 epochs. For PDO, we apply early stopping with a patience of 20 epochs and a minimum validation-loss improvement of \(10^{-4}\). In every experiment, model selection is performed using validation loss, and the checkpoint achieving the lowest validation loss is subsequently evaluated on the held-out test set. PDO and Charville populations are embedded into 32-dimensional latent spaces, whereas UPMC populations are embedded into a 16-dimensional latent space. Unless otherwise specified, we use global mean pooling, $J=2$ wavelet scales, and a PhenoGS regularization weight of \(0.1\). For PDO, we use \(K=1\); for the spatial Charville and UPMC datasets, we use \(K=2\). The conditional latent-flow generator is optimized using Adam optimizer with a learning rate of \(10^{-3}\) for 30,000 epochs, while the point-cloud decoder is trained for 400 epochs. For the interpolation experiment, we constructed a \(20\)-nearest-neighbor graph over the learned population-level latent representations and identified a shortest path between selected endpoint populations. All representation-learning and generation experiments are performed on a Nvidia H200 GPU with four CPU workers.

\subsection{Data Splits}\label{appsub:datasplits}
For PDO, patients 75 and 99 are held out for validation and testing,
respectively, and the remaining patients are used for training. The same
patient-based split is used when estimating standardization parameters. For
Charville, patients c001--c003 are used for training, while the populations
from c004 are randomly divided equally between validation and testing. For
UPMC, patients c001--c004 and c007 are used for training. The combined
populations from c005 and c006 are shuffled and divided equally between
validation and testing. All randomized operations use seed 1302 unless stated otherwise. The PDO path connected a DMSO control population to an F-treated population from patient 21 and contained three held-out intermediate populations corresponding to L-, S-, and S-treated cultures. The Charville path connected non-recurrent and recurrent populations from patient c001 and contained two held-out intermediate populations. The UPMC path connected non-recurrent and recurrent populations from patient c005 and contained three held-out intermediate populations; because the graph was constructed over the complete cohort, two intermediate populations belonged to patients c006 and c007

\subsection{Metrics}\label{appsub:metrics}

We evaluate generated point clouds using Earth Mover's Distance
(EMD, or $W_1$), the 2-Wasserstein distance ($W_2$), and maximum
mean discrepancy (MMD). We selected these metrics because they provide
complementary assessments of point-cloud generation quality. Both
Wasserstein distances compare populations directly in marker space while
accounting for the complete empirical distributions rather than requiring
predefined cell-to-cell correspondences. In particular, EMD measures the
minimum average displacement required to transport the generated population
onto the real population and is comparatively robust to isolated large
errors. By contrast, $W_2$ uses a squared ground cost and therefore penalizes
large marker-space discrepancies more strongly. MMD complements these
transport-based metrics by comparing distributions through kernel similarities,
allowing it to detect nonlinear differences in their overall structure. Using
multiple kernel bandwidths makes the MMD evaluation sensitive to discrepancies
at several spatial scales.

Let $\widehat{\mathcal{X}}$ and $\mathcal{X}$ denote generated and real
point clouds, respectively. EMD is the optimal transport cost under the
Euclidean ground distance:
\begin{equation}
W_1(\widehat{\mathcal{X}},\mathcal{X})
=
\min_{\pi\in\Pi(\widehat{\mathcal{X}},\mathcal{X})}
\sum_{i,j}\pi_{ij}
\left\lVert \widehat{\mathbf{x}}_i-\mathbf{x}_j \right\rVert_2,
\end{equation}
where $\Pi(\widehat{\mathcal{X}},\mathcal{X})$ is the set of couplings
between the uniform empirical measures defined by the two point clouds.

We compute $W_2$ using the squared Euclidean ground cost and report the
square root of the resulting optimal transport cost:
\begin{equation}
W_2(\widehat{\mathcal{X}},\mathcal{X})
=
\left(
\min_{\pi\in\Pi(\widehat{\mathcal{X}},\mathcal{X})}
\sum_{i,j}\pi_{ij}
\left\lVert \widehat{\mathbf{x}}_i-\mathbf{x}_j \right\rVert_2^2
\right)^{1/2}.
\end{equation}

MMD compares the generated and real empirical distributions in a
reproducing-kernel Hilbert space. For generated cells
$\widehat{\mathcal{X}}=\{\widehat{\mathbf{x}}_i\}_{i=1}^{n}$ and real
cells $\mathcal{X}=\{\mathbf{x}_j\}_{j=1}^{m}$, we compute the biased
empirical squared MMD:
\begin{align}
\operatorname{MMD}_{k}^{2}
(\widehat{\mathcal{X}},\mathcal{X})
={}&
\frac{1}{n^{2}}
\sum_{i,i'=1}^{n}
k(\widehat{\mathbf{x}}_i,\widehat{\mathbf{x}}_{i'})
+
\frac{1}{m^{2}}
\sum_{j,j'=1}^{m}
k(\mathbf{x}_j,\mathbf{x}_{j'})
\nonumber\\
&-
\frac{2}{nm}
\sum_{i=1}^{n}\sum_{j=1}^{m}
k(\widehat{\mathbf{x}}_i,\mathbf{x}_j),
\end{align}
where the radial basis function kernel is
\begin{equation}
k_{\gamma}(\mathbf{x},\mathbf{y})
=
\exp\left(
-\gamma\left\lVert\mathbf{x}-\mathbf{y}\right\rVert_2^2
\right).
\end{equation}
We report the mean of
$\operatorname{MMD}_{k_\gamma}^{2}$ over
$\gamma\in\{2,1,0.5,0.1,0.01,0.005\}$, thereby evaluating agreement
across multiple kernel scales. Lower values of all three metrics indicate
greater agreement between the generated and real cell distributions.

\section{Computational Complexity} \label{app:complexity}
Let $M$ denote the number of point clouds, $D$ is the latent dimension, and $n^{(m)}$ the number of points in the point cloud $\mathcal{X}^{(m)}$. Training $\mathcal{L}_{GS}$ over a minibatch of $B$ point clouds costs $O(B^2D)$, while the target distance matrix $W$ is precomputed once at $O(M^2)$ memory and cost dominated by diffusion over a shared point-level graph. Training the generation head costs $O((n^{(m)})^2D)$ per point cloud under a fixed-iteration Sinkhorn approximation. Latent flow matching operates on latents $\mathbf{z}^{(m)}$ so training over $M$ point clouds for $E$ epochs costs $O(EMV)$, where $V$ is the cost of one forward pass through the velocity network, and sampling $J$ latents with $S$ Euler steps costs $O(SJV)$, both independent of $n^{(m)}$. Decoding the point clouds then adds $O(J\bar{n}V_G)$, where $\bar{n}$ is the mean number of points across $J$ generated point clouds and $V_G$ is the cost of one forward pass through the generation head. Interpolation builds a $k$-nearest neighbor graph over the $M$ latents and runs Dijkstra's algorithm at $O((M+Mk)\log M)$ per query. 

\section{Additional Results}\label{sec:appendix-results}

\subsection{Additional Phenoscapes}
\begin{figure}[t]
  \centering
  \includegraphics[width=\textwidth]{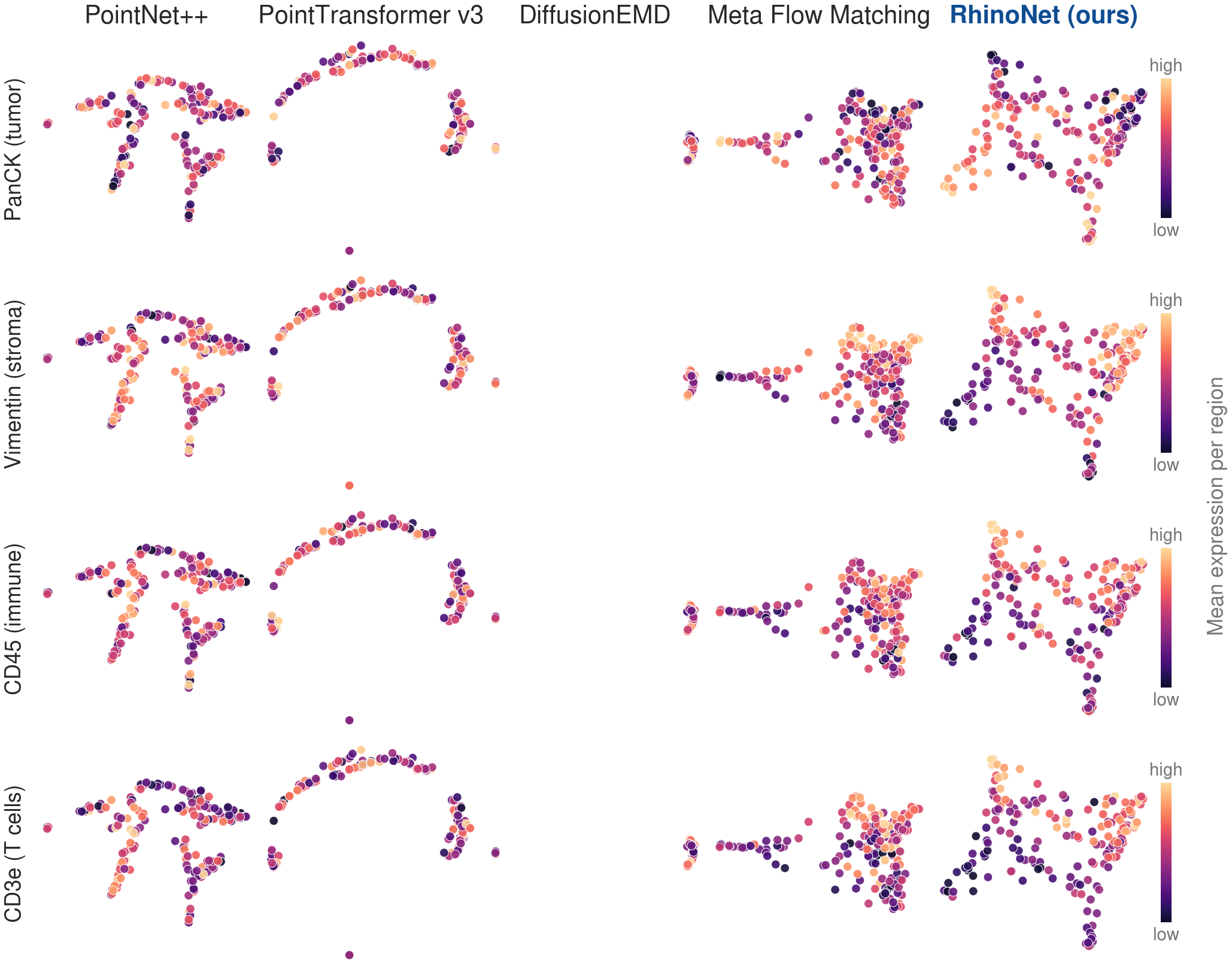}
  \caption{\textbf{RhinoNet organises tissue regions along a tumour--stroma axis.} Phenoscapes of 186 CODEX tissue regions from the Charville cohort. Each dot is one region, placed by 2D PHATE ($k=5$) of that method's latent space and coloured by the region's mean expression of a marker for tumour (PanCK), stroma (Vimentin), immune cells (CD45) or T cells (CD3e). In RhinoNet's phenoscape, tumour-rich and stroma- and immune-rich regions lie at opposite ends of a continuous axis, while the baselines show no consistent gradient. At $k=5$ the DiffusionEMD neighbour graph splits by slide, so PHATE places the two components far apart.}
  \label{fig:charville_phenoscape}
\end{figure}

\begin{figure}[t]
      \centering
      \includegraphics[width=\textwidth]{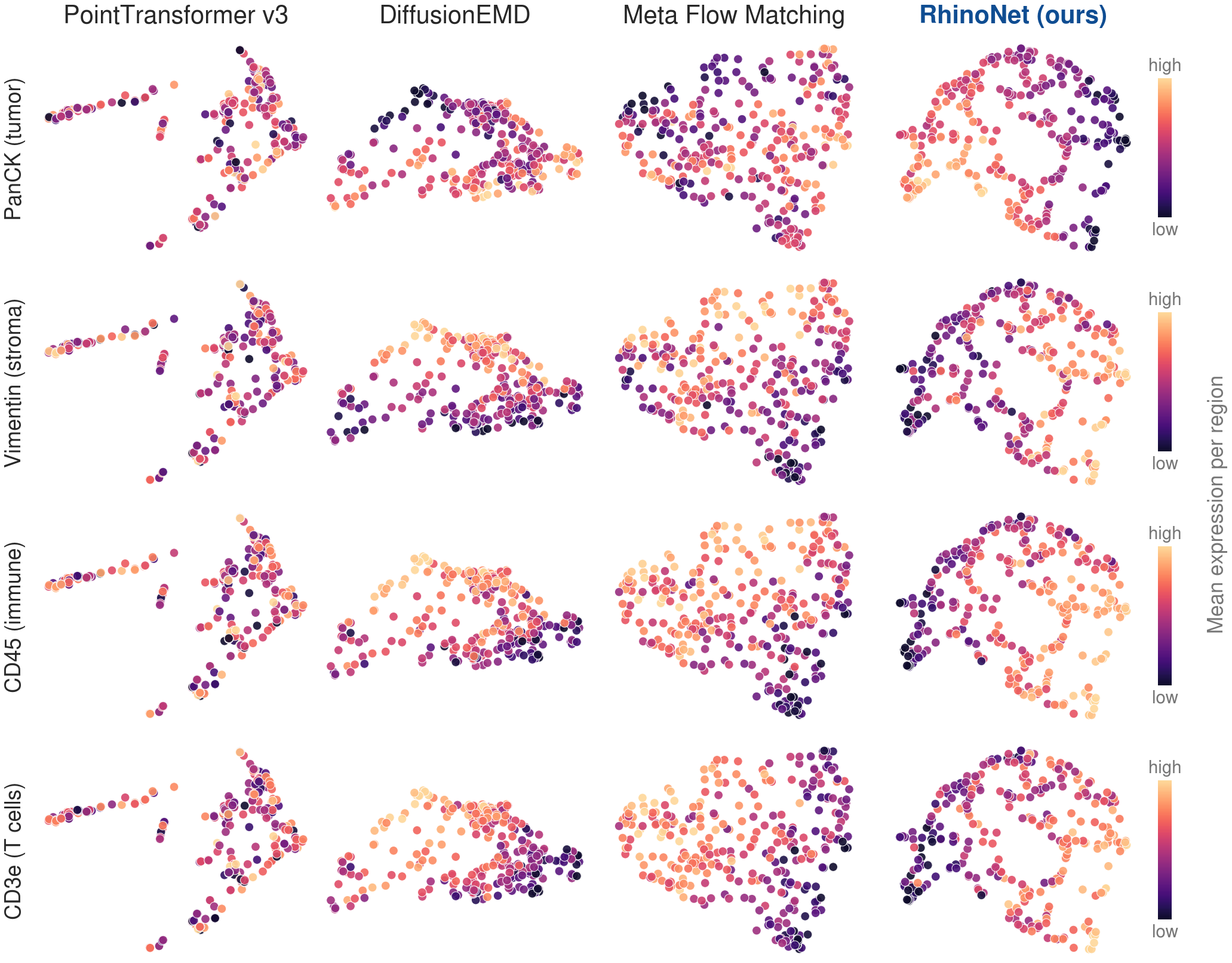}
      \caption{\textbf{UPMC phenoscapes.} 308 CODEX tissue regions from the UPMC cohort, one dot per region, placed by 2D PHATE of each method's latent space and coloured by the region's mean expression of markers for tumour (PanCK), stroma (Vimentin), immune cells (CD45) and T cells (CD3e). RhinoNet orders regions from tumour-rich (high PanCK) to stroma- and immune-rich (high Vimentin, CD45 and CD3e).}
      \label{fig:upmc_phenoscape}
\end{figure}

We next construct phenoscapes of spatial proteomics data from the Charville cohort~\citep{space-gm}, in which each point cloud is a tissue region imaged by CODEX, and each cell is described by 40 protein markers. We colour each region by the mean expression of markers for the main tissue compartments: tumour (PanCK), stroma (Vimentin), immune cells (CD45) and T cells (CD3e). Figure~\ref{fig:charville_phenoscape} shows that RhinoNet arranges the regions along a continuous axis from tumour-dominated regions, with high PanCK, to regions rich in stroma and immune infiltrate, with high Vimentin, CD45 and CD3e. The marker gradients change smoothly across the phenoscape, even though RhinoNet is never given marker-level supervision. In the PointNet++, PointTransformer and Meta Flow Matching phenoscapes, regions with high and low expression are interleaved, with no consistent gradient. DiffusionEMD instead separates regions by slide rather than by tissue composition, which suggests that it is dominated by a batch effect.   On the UPMC cohort, RhinoNet again orders regions along a tumour--immune axis (Figure~\ref{fig:upmc_phenoscape}), and it keeps the tumour marker smoothest across latent neighbourhoods of all methods tested. Across all 22 markers, DiffusionEMD, which operates directly on optimal transport between cell distributions, gives the smoothest gradients, while PointTransformer again collapses the cohort without a consistent gradient.

\subsection{Hodge Laplacian Ablation}\label{app:hodge-laplacian-ablation}
\begin{figure}[h]
      \centering
      \includegraphics[width=0.6\textwidth]{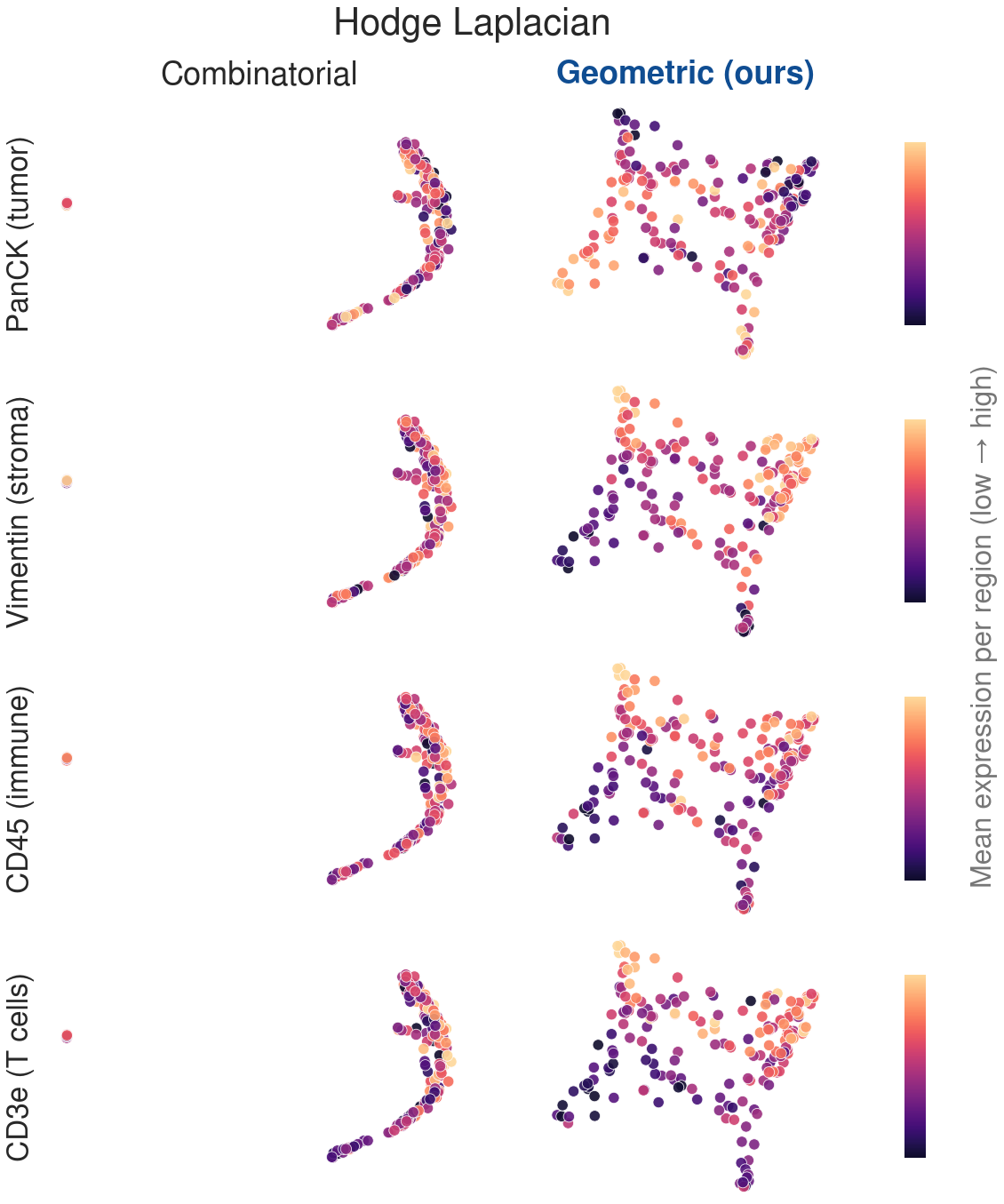}
      \caption{Hodge Laplacian ablation on Charville, coloured by PanCK, Vimentin, CD45 and CD3e.}
      \label{fig:laplacian_ablation_full}
  \end{figure}

Figure~\ref{fig:laplacian_ablation_full} compares Charville phenoscapes from two RhinoNet models that differ only in the Hodge Laplacian used by the simplicial wavelet transform: the combinatorial Laplacian (left) and our geometric Laplacian (right). Both models use $K=2$, the same training data, seed and hyperparameters, and are visualised with PHATE on the same 186 tissue regions. Each region is coloured by its mean expression of markers for tumour (PanCK), stroma (Vimentin), immune cells (CD45) and T cells (CD3e). With the geometric Laplacian, the phenoscape spreads regions over two dimensions, and stroma- and immune-rich regions separate from tumour-rich ones. With the combinatorial Laplacian, most regions collapse onto a single arc, and 74\% of the latent variance lies along one direction.

\subsection{Interpolation Paths}\label{app:interpolation-paths}
\begin{figure}[t]
    \centering
    \includegraphics[width=\textwidth]{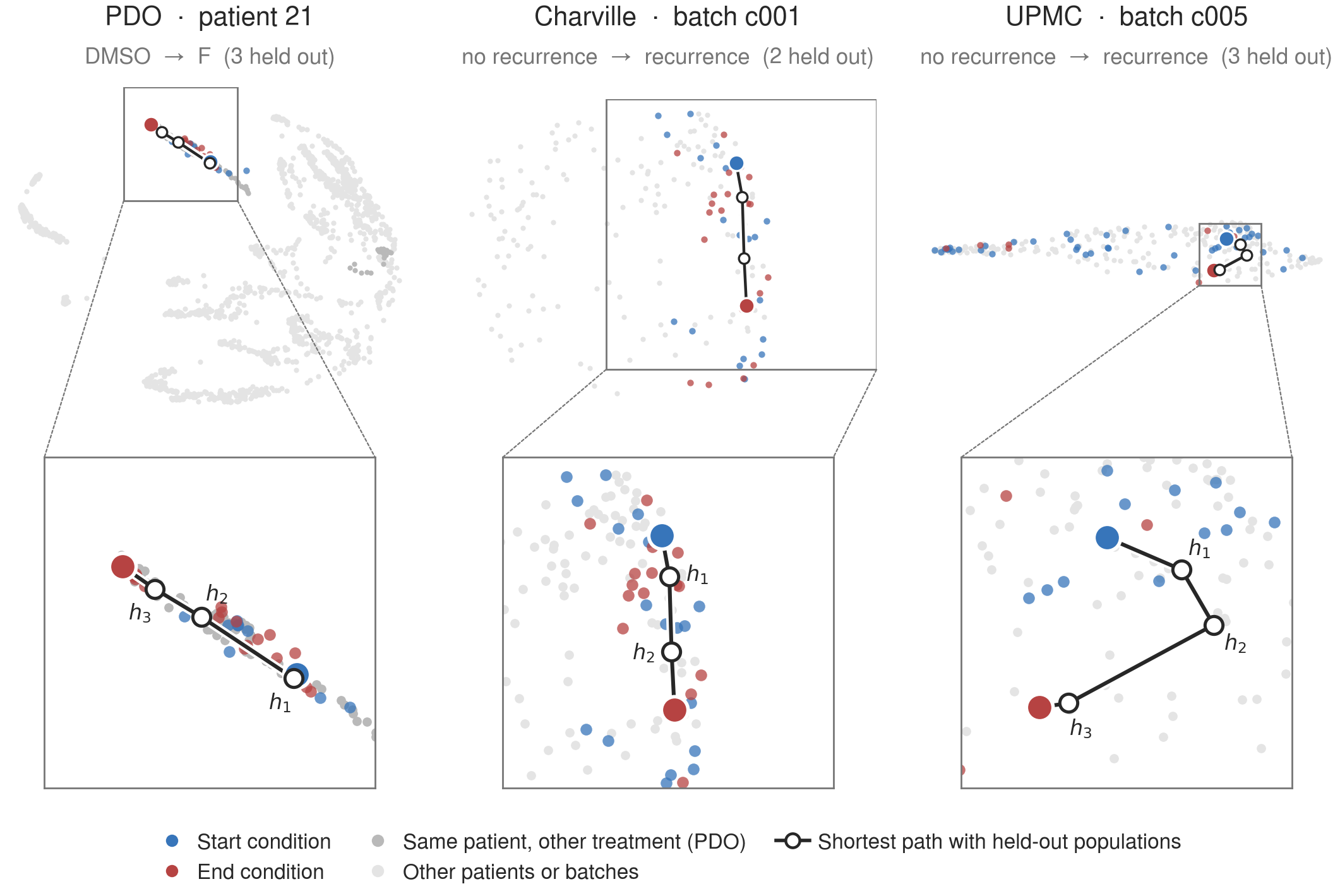}
    \caption{\textbf{Interpolation paths through RhinoNet's phenoscape.} \textbf{Top:} 2D PHATE of RhinoNet's latent space for PDO, Charville and UPMC, one dot per point cloud. Point clouds of the interpolated patient are coloured by condition (blue: start, red: end), the same patient's other treatments are dark grey (PDO only), and all other patients are light grey. The box marks the region enlarged below. \textbf{Bottom:} the shortest path through the $k$-NN graph over latents (black line) between two endpoints of different conditions (filled circles). Held-out intermediate point clouds $h_1,\dots,h_n$ (open circles) are numbered in path order. Both rows use equal axis scales, so distances in the enlarged panels are not distorted.}
    \label{fig:interpolation_paths}
  \end{figure}

 Figure~\ref{fig:interpolation_paths} shows the discrete geodesics used in the interpolation experiment (Section~\ref{sec:datagen}, Figure~\ref{fig:interpolation}). Each dot is one point cloud, placed by 2D PHATE of RhinoNet's latent space. For each cohort, the path runs through a symmetrized 20-nearest-neighbour graph over the latents, from a point cloud of the start condition to the nearest point cloud of a different condition from the same patient or batch. The intermediate point clouds on the path are held out from decoder training and reconstructed from their latent codes. On PDO, the path from a DMSO culture to a treatment-F culture of patient 21 passes through three held-out cultures of the same patient, treated with LGK-974 (L) and SN-38 (S). On Charville, the path runs from a non-recurrent to a recurrent region of batch c001, and its two held-out regions come from the same batch: the first is non-recurrent and the second is recurrent. On UPMC, the path joins a non-recurrent and a recurrent region of batch c005. Its three held-out regions come from batches c006, c007 and c005, so the geodesic passes through neighbouring batches' samples. 

\subsection{Simplex Order Ablations}\label{app:k_ablation}
\begin{figure}[t]
      \centering
      \begin{subfigure}[t]{0.49\textwidth}
          \centering
          \includegraphics[width=\linewidth]{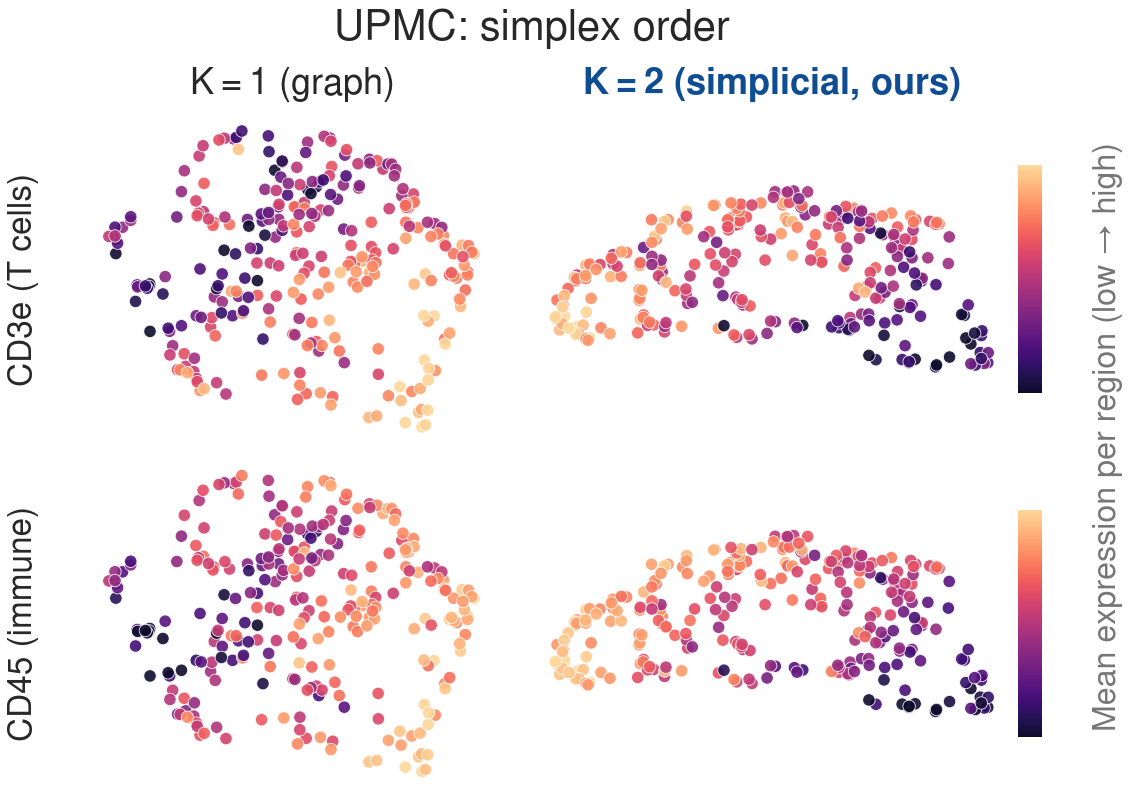}
          \caption{UPMC}
          \label{fig:k_ablation_upmc}
      \end{subfigure}\hfill
      \begin{subfigure}[t]{0.49\textwidth}
          \centering
          \includegraphics[width=\linewidth]{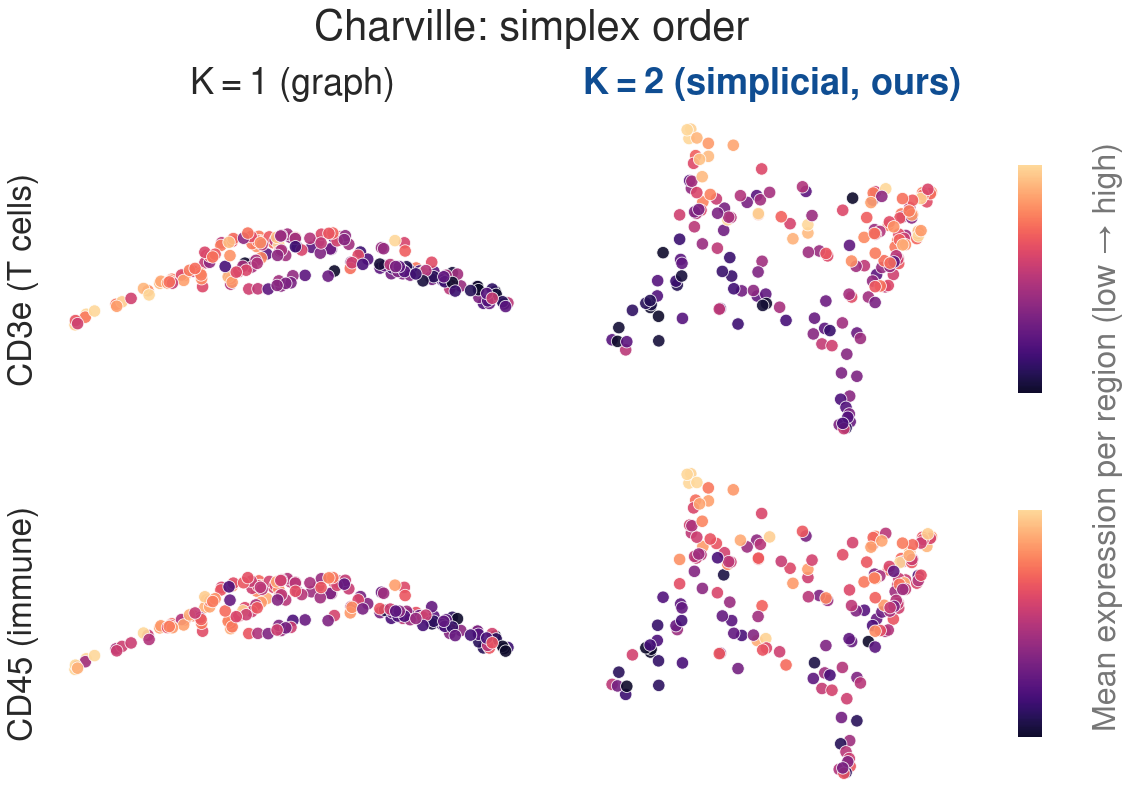}
          \caption{Charville}
          \label{fig:k_ablation_charville}
      \end{subfigure}
      \caption{\textbf{Ablation of the simplex order $K$.} Phenoscapes from RhinoNet with $K=1$ (graph: nodes and edges) or $K=2$ (simplicial: nodes, edges and triangles), coloured by mean CD3e and CD45 expression per tissue region. On UPMC, $K=2$ orders regions along a single axis from immune-rich to immune-poor. On Charville, both models separate immune-rich from immune-poor regions.}
      \label{fig:k_ablation}
  \end{figure}

We next ablate the order of the simplicial complex, comparing $K=2$, which adds triangles to the complex, with $K=1$, which reduces the encoder to a graph over cells. On both spatial cohorts, $K=2$ fits the geodesic Sinkhorn distances between tissue regions more closely and reconstructs the wavelet features more accurately. The resulting phenoscapes (Figure~\ref{fig:k_ablation}) show that on UPMC, $K=2$ arranges regions along a continuous axis of immune infiltration, with smoother CD3e and CD45 gradients across latent neighbourhoods than $K=1$. On Charville, both models recover the immune gradient, with $K=1$ producing a smoother one-dimensional ordering, while $K=2$ gives a closer fit to the transport distances. All markers are shown in Appendix~\ref{app:k_ablation}.

\end{document}